\documentclass{article}

\PassOptionsToPackage{numbers, compress}{natbib}

\usepackage[preprint]{neurips_2025}

\usepackage[dvipsnames]{xcolor}

\usepackage[utf8]{inputenc} 
\usepackage[T1]{fontenc}    
\usepackage{hyperref}       
\usepackage{url}            
\usepackage{booktabs}       
\usepackage{amsfonts}       
\usepackage{nicefrac}       
\usepackage{microtype}      
\usepackage{mathtools}
\usepackage{graphicx}
\usepackage{subfig}
\usepackage{wrapfig}
\usepackage{multirow}
\usepackage{booktabs}       
\usepackage{bbding}

\usepackage{color, colortbl}

\definecolor{GM}{cmyk}{0.1093, 0.0526, 0, 0.0314}
\definecolor{DI}{cmyk}{0, 0.0956, 0.1474, 0.0157}
\definecolor{Compositor}{cmyk}{0, 0.045, 0.2, 0.01}

\title{\textit{DetailFusion}: A Dual-branch Framework with \\
       Detail Enhancement for Composed Image Retrieval}

\author{%
Yuxin~Yang$^{1,2,3}$ \quad Yinan~Zhou$^{4,5}$ \quad Yuxin~Chen$^5$ \quad Ziqi~Zhang$^{1,2,6}$ \quad Zongyang~Ma$^{1,2,3}$ \\
\textbf{Chunfeng~Yuan}$^{1,2}$\thanks{Corresponding author} \quad \textbf{Bing~Li}$^{1,2,6}$ \quad \textbf{Lin~Song}$^5$ \quad \textbf{Jun~Gao}$^7$ \quad \textbf{Peng~Li}$^8$ \quad \textbf{Weiming~Hu}$^{1,2,3,9}$ \\
$^1$Beijing Key Laboratory of Super Intelligent Security of Multi-Modal Information, CASIA \\
$^2$State Key Laboratory of Multimodal Artificial Intelligence Systems, CASIA \\
$^3$School of Artificial Intelligence, University of Chinese Academy of Sciences \\
$^4$Xi'an Jiaotong University \quad $^5$ARC Lab, Tencent PCG \\
$^6$PeopleAI Inc. \quad $^7$HelloGroup Inc. \quad $^8$Xiaomi Group \\
$^9$School of Information Science and Technology, ShanghaiTech University \\
\texttt{\{yangyuxin2023, mazongyang2020\}@ia.ac.cn, zyn13572297710@stu.xjtu.edu.cn} \\
\texttt{\{ziqi.zhang, cfyuan, bli, wmhu\}@nlpr.ia.ac.cn}
}

\begin{document}

\maketitle

\begin{abstract}

Composed Image Retrieval (CIR) aims to retrieve target images from a gallery based on a reference image and modification text as a combined query.
Recent approaches focus on balancing global information from two modalities and encode the query into a unified feature for retrieval. However, due to insufficient attention to fine-grained details, these coarse fusion methods often struggle with handling subtle visual alterations or intricate textual instructions.
In this work, we propose \textbf{\textit{DetailFusion}}, a novel dual-branch framework that effectively coordinates information across global and detailed granularities, thereby enabling detail-enhanced CIR.
Our approach leverages atomic detail variation priors derived from an image editing dataset, supplemented by a detail-oriented optimization strategy to develop a \textit{Detail-oriented Inference Branch}. Furthermore, we design an \textit{Adaptive Feature Compositor} that dynamically fuses global and detailed features based on fine-grained information of each unique multimodal query.
Extensive experiments and ablation analyses not only demonstrate that our method achieves state-of-the-art performance on both CIRR and FashionIQ datasets but also validate the effectiveness and cross-domain adaptability of detail enhancement for CIR.
The code will be publicly available at \url{https://github.com/HaHaJun1101/DetailFusion}.

\end{abstract}

\section{Introduction}
\label{sec:introduction}

Composed Image Retrieval (CIR)~\cite{tirg, cirr} is an advanced image retrieval task that aims to retrieve target images from a gallery by utilizing a dual-modality query consisting of a reference image and a modification text. Unlike traditional image retrieval that relies solely on either visual~\cite{visual_1, visual_2} or textual~\cite{textual_1, textual_2, textual_3} information, CIR leverages the synergy between these two modalities to provide more precise and contextually relevant search results that satisfy user intent, thereby expanding its application scenarios to a wider range, including e-commerce and interactive search systems.

\begin{figure}[t]
\vspace{-\baselineskip}
\begin{center}
\centerline{\includegraphics[width=1.01\linewidth, trim=0.6cm 0.55cm 0.45cm 0.05cm, clip]{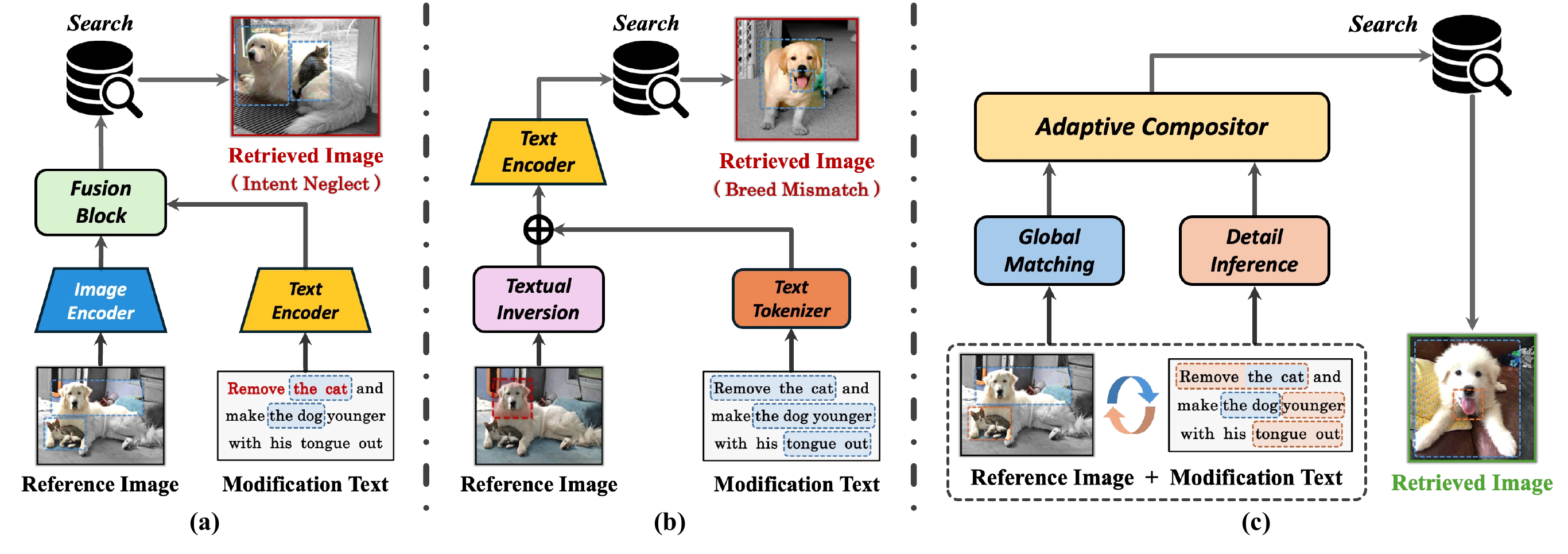}}
\vspace{-0.2\baselineskip}
\caption{Workflows of existing supervised CIR methods and Ours: (a) Late fusion, (b) Textual inversion, and (c) Our proposed \textbf{\textit{DetailFusion}}. Due to the lack of dedicated modules for fine-grained details, the first two prevalent CIR methods struggle to \textit{perceive fine-grained details} in the reference image while \textit{processing complex requirements} in the modification text. (a) Fails to implement the textual transformation \textit{`Remove'} and misses the requirement \textit{`tongue out'}. (b) Overlooks implicit visual detail information, failing to preserve \textit{the dog's breed}. (c) Our approach combines global semantic understanding with fine-grained detail interaction, resulting in accurate retrieval outcomes.}
\label{fig:workflows}
\end{center}
\vspace{-1.5\baselineskip}
\end{figure}

Many existing supervised CIR methods leverage Vision-Language Pre-training (VLP) models~\cite{clip, blip, blip2} to perform single-modal encoding for the reference images and modification texts separately, and subsequently employ various fusion strategies to integrate global semantic information from both modalities into a unified feature for the retrieval process. These methods focus on balancing and coordinating information from the two modalities to minimize information loss during the fusion process. However, as illustrated in Fig.~\ref{fig:workflows}(a) and Fig.~\ref{fig:workflows}(b), these coarse fusion methods often encounter two typical problems: (1) \textbf{\textit{Insufficient visual detail perception}}: The inability to \textit{perceive fine-grained details} in the reference image, leading to failures in retrieving the target image with \textit{subtle alterations}. (2) \textbf{\textit{Insufficient textual detail processing}}: When handling \textit{complex modification requirements}, certain \textit{detailed atomic transformations} may be overlooked. We argue that the root cause of these drawbacks lies in the model's lack of targeted training to perceive and process fine-grained details.

In this work, we overcome the shortcomings of existing supervised CIR methods through detail enhancement. Throughout this process, we encounter two significant challenges: (1) \textbf{\textit{Efficient detail-oriented training}}: Enabling the model to focus on visual details in both reference and target images while accurately understanding and executing textual atomic transformations in the modification text. (2) \textbf{\textit{Coordination of multi-granularity information}}: Enabling the model to handle both overall alterations and detail modifications during the encoding process. To address the above challenges, we propose \textbf{\textit{DetailFusion}}, an innovative dual-branch framework designed to enhance the perception and processing of fine-grained details while maintaining the capability to extract global semantics.

As illustrated in Fig.~\ref{fig:workflows}(c), our proposed \textbf{\textit{DetailFusion}} treats the reference image and the modification text as a unified input for multimodal fusion encoding, promoting fine-grained interactions between the two modalities. For a multimodal query, the \textit{\textbf{G}lobal Feature \textbf{M}atching (GM) Branch} captures comprehensive semantic information, while the \textit{\textbf{D}etail-oriented \textbf{I}nference (DI) Branch} focuses on fine-grained details for reasoning-based validation. The \textit{Adaptive Feature \textbf{Compositor}} then dynamically fuses global and detailed features from the corresponding branch for each unique query. This strategy of separately managing different granularities information enables more precise retrieval outcomes.

To optimize the performance of each module, we design a \textbf{\textit{three-stage training strategy}} for the entire framework. The process can be summarized as follows: \textbf{\textit{Detail}} — Developing a branch dedicated to fine-grained details. \textbf{\textit{Fusion}} — Dynamically fusing the global and detailed features in a coarse-to-fine manner. 
Specifically, leveraging a hybrid-modal encoder~\cite{blip2}, we utilize triplet data with subtle image modifications derived from a large scale image editing dataset~\cite{ipr2pr} to pre-train the DI branch, complemented by a \textit{detail-oriented optimization strategy} that prevents the model from relying on image similarity shortcuts. This guides the DI branch to disregard highly similar yet irrelevant visual information while concentrating on variations in details. Subsequently, we jointly fine-tune the DI and GM branches on CIR datasets to achieve coarse alignment with the encoded features of the target image. Finally, we train the Compositor from scratch to enable dynamic and refined fusion of features from both branches, while injecting fine-grained information to enhance the final representation.

In summary, the main contributions of this work are listed as follows:
\begin{itemize}
\item
    We are the first to apply an image editing dataset for training in the CIR task, achieving cross-task data adaptation and application in vision-language domains. By employing a detail-oriented optimization strategy, we address the challenges posed by the high similarity between reference and target images, designing a dedicated branch for fine-grained details.
\item
    We propose \textbf{\textit{DetailFusion}}, a powerful framework for CIR that excels in perceiving and processing fine-grained details. Its dual-branch architecture effectively processes information across multiple levels of granularity, while the Compositor dynamically enhance the final representation for each query. Additionally, we develop a three-stage training strategy that optimizes the performance of each module within the framework in a coarse-to-fine manner.
\item
    Extensive experiments on the CIRR and FashionIQ datasets demonstrate that \textbf{\textit{DetailFusion}} outperforms the state-of-the-art CIR methods. Furthermore, we validate the effectiveness of detail enhancement and cross-domain adaptability of enhanced detail capabilities for CIR.
\end{itemize}

\section{Related Work}
\label{sec:related_work}

\textbf{Composed Image Retrieval.}
The prevailing supervised CIR methods typically leverage VLP~models for foundational encoding, subsequently employing various strategies to tailor them for the CIR task.
These approaches can be classified into two categories: fusion-based and textual inversion-based.
The \textit{late-fusion} approach~\cite{artemis, composeae, clip4cir, spirit, cqbir} integrates features from the reference image and the modification text after independent encoding.
In contrast, \textit{early-fusion} approaches~\cite{dataroaming, reranking, dqu} merge image and text embeddings during encoding through cross-attention layers in a hybrid-modal encoder, enabling more nuanced token-patch level interactions.
However, without dedicated detail‑oriented training, these fusion-based approaches often struggle to capture fine‑grained visual details or accurately execute textual atomic transformations.
The \textit{textual inversion} approach \cite{palavra, clip_inversion} generates a pseudo-word embedding~\cite{pic2word, searle} or a sentence-level prompt~\cite{sprc} based on the reference image, which is then concatenated with the modification text for text-to-image retrieval. While this approach accommodates more complex textual requirements, it often suffers from inadequate image representation, leading to the loss of implicit visual information.
In this work, we address the above challenges from the perspective of detail enhancement. Building upon the early-fusion approach, we aim to exploit the model's capacity for perceiving and processing fine-grained details to improve the CIR performance.

\textbf{CIR Pre-training.} Recent approaches have leveraged automatically annotated triplets to pre-train the CIR models.
LaSCo~\cite{dataroaming}, CoVR~\cite{covr}, and MagicLens~\cite{magiclens} filter relevant image or video pairs from datasets~\cite{vqa2, webvid} or web sources and fine-tune Large Language Models (LLMs)~\cite{llama, gpt3} to generate modification text.
Alternatively, InstructPix2Pix~\cite{ipr2pr} and CompoDiff~\cite{compodiff} generate relative image pairs using Stable Diffusion~\cite{stable_diffusion} based on manually annotated or generated modification text.
However, these automatic annotation approaches often lead to erroneous or trivial triplets, which limit their effectiveness for pre-training CIR models.
In this work, to fully leverage the high similarity of image pairs and the textual atomic transformations in image editing datasets, we focus on pre-training specific modules for detail enhancement, rather than pre-training the entire CIR model directly.

\section{Method}
\label{sec:method}

\subsection{Problem Statement}
\label{subsec:problem_statement}

Assume $ Q = \{I_r, T_m\} $ denotes the input dual-modality query, where $I_r$ is the reference image, and $T_m$ is the modification text. The objective of Composed Image Retrieval (CIR) is to retrieve the target image $I_t$ from the image gallery $\mathbf{G}$ that best matches the query $Q$:
\begin{equation}
    I_t = \arg\max_{I \in \mathbf{G}} Sim(\mathcal{V_{\,I}}(I), \mathcal{V_{M}}(Q)),
\label{eq:1}
\end{equation}
where $Sim(\cdot)$ represents the cosine similarity between the encoded feature vectors, $\mathcal{V_{\,I}}(\cdot)$ and $\mathcal{V_{M}}(\cdot)$ denote the feature encoding process for the image and multimodal query respectively, and $I$ denotes a single image in the image gallery.

\begin{figure}[t]
\vspace{-0.8\baselineskip}
\begin{center}
\centerline{\includegraphics[width=\linewidth]{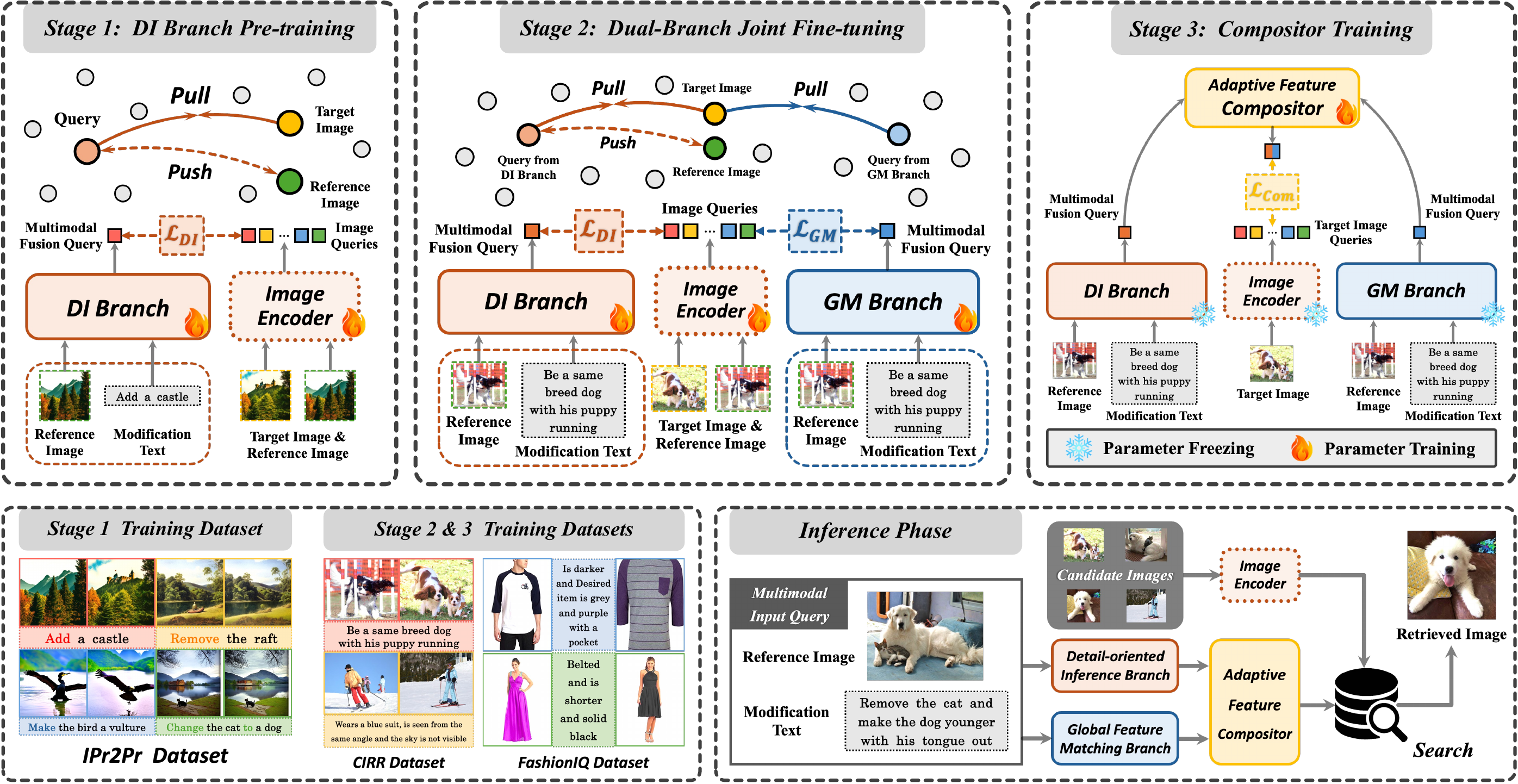}}
\vspace{0.1\baselineskip}
\caption{Overall pipeline of our proposed \textbf{\textit{DetailFusion}}. The upper section illustrates the \textit{training phase}, while the lower-left displays the \textit{datasets} utilized at each training stage, and the lower-right illustrates the \textit{inference phase}. During the \textit{training phase}, the DI branch is first pre-trained on the IPr2Pr dataset~\cite{ipr2pr}, followed by joint fine-tuning of both the DI and GM branches on CIR datasets. Finally, the parameters of both branches are frozen, and the Compositor is trained from scratch on the corresponding CIR datasets. During the \textit{inference phase}, the multimodal query is encoded through the DI and GM branches to extract fine-grained detail and global semantic features, respectively. These features are subsequently fused and enhanced by the Compositor to produce the final representation.}
\label{fig:training_and_inference}
\end{center}
\vspace{-1.5\baselineskip}
\end{figure}

\subsection{Method Overview}
\label{subsec:method_overview}


As illustrated in Fig.~\ref{fig:training_and_inference}, our proposed \textbf{\textit{DetailFusion}} framework comprises three key modules: a \textit{\textbf{D}etail-oriented \textbf{I}nference (DI) Branch} that focuses on fine-grained detail interactions within the query, a \textit{\textbf{G}lobal Feature \textbf{M}atching (GM) Branch} that captures comprehensive semantic information, and an \textit{Adaptive Feature \textbf{Compositor}} that dynamically fuses the global and detailed features, injecting fine-grained auxiliary information to enhance the final representation for each multimodal query.

We employ a three-stage training strategy to optimize the framework effectively: In the first stage, the DI branch is pre-trained using an image editing dataset. We utilize reference images as hard negative samples, guiding the DI branch to perceive critical variations in fine-grained details.
In the second stage, both the DI and GM branches are fine-tuned on CIR datasets to achieve a coarse alignment with the encoded feature of the target image. In the third stage, the Compositor is trained from scratch to enable fine alignment between the fused global and detailed feature and the target image feature. 

Additionally, we reuse the hybrid-modal encoder structure of the DI and GM branches for efficiency. Within the Compositor, we incorporate a dual-branch cross-attention mechanism that facilitates information interaction across different granularities, while injecting fine-grained tokens information, thereby promoting the subsequent feature fusion process and enhancing the final representation.

\subsection{Three-Stage Training Strategy}
\label{subsec:three-stage_training_strategy}

\textbf{Pre-training of \textit{DI} Branch.}
Before training the entire framework on CIR datasets, we first pre-train the DI branch using a large-scale image editing dataset, IPr2Pr~\cite{ipr2pr}. This dataset shares the same triplet data structure as CIR datasets but differs in two aspects: (1) \textit{High Similarity Between Reference and Target Images}: The images in IPr2Pr are specially generated for image editing tasks, resulting in triplets which the reference and target images share nearly identical overall semantic content, with only subtle variations in details. (2) \textit{Textual Atomic Transformations}: The modification texts in IPr2Pr are concise and only describe single-step operations commonly encountered in image editing tasks. 

To prevent the DI branch from relying on image similarity shortcuts and enable it effectively perceive subtle visual transformations while comprehending the essence of textual atomic modifications, we propose a detail-oriented optimization strategy.
Building upon the batch-based contrastive learning, we introduce the reference image itself as a hard negative for the corresponding query during training.
The detail-oriented inference loss function is formulated as follows: \par
\vspace{-\baselineskip}
\begin{small}
    \begin{equation}
        \mathcal{L}_{D \! I} = -\frac{1}{\mathcal{|B|}} \sum_{i=1}^{\mathcal{|B|}} \log 
        \frac{\mathbb{S}(\mathcal{D}(Q^{(i)}), \mathcal{D}(I^{(i)}_{t}))}
        { \sum_{j=1}^{\mathcal{|B|}} \mathbb{S}(\mathcal{D}(Q^{(i)}), \mathcal{D}(I^{(j)}_{t})
        \bigoplus \mathcal{D}(I^{(j)}_{r})) },
    \label{eq:2}
    \end{equation}
\vskip 0.02in
\end{small}%
where { \small $\mathbb{S}(\cdot) \coloneqq \exp(Sim( (\cdot) / \tau)$ } and { \small $\mathbb{S}(A, B \bigoplus C) \coloneqq \mathbb{S}(A, \! B) + \mathbb{S}(A, \! C)$ }, $\mathcal{B}$ denotes the training batch, $Q^{(i)} = \{I^{(i)}_r, T^{(i)}_m\}$ and $I^{(i)}_t$ denotes the $i$-th dual-modality query and target image in $\mathcal{B}$, respectively. $\mathcal{D}(\cdot)$ denotes the feature encoding process of the DI branch, and ${\tau}$ is a temperature hyper-parameter.

This strategy forces the DI branch to ignore highly similar but irrelevant visual information and focus on subtle variations between image pairs. As a result, the model learns to establish precise correlations between decoupled textual atomic transformations and their corresponding visual manifestations.

\textbf{Joint Fine-tuning of Dual-Branch.}
To fully leverage the unique capabilities of both branches, we employ distinct optimization strategies during fine-tuning on CIR datasets. For the DI branch, we retain the loss function $\mathcal{L}_{D\!I}$ to ensure that the detail-related capabilities gained during pre-training are effectively transferred to the CIR task. In contrast, the GM branch utilizes batch-based contrastive learning, which is sufficient for extracting and integrating global semantics: \par
\vspace{-0.8\baselineskip}
\begin{small}
    \begin{equation}
        \mathcal{L}_{G \! M} = -\frac{1}{\mathcal{|B|}} \sum_{i=1}^{\mathcal{|B|}} \log
        \frac{\mathbb{S}(\mathcal{G}(Q^{(i)}), \mathcal{D}(I^{(i)}_{t}))}
        { \sum_{j=1}^{\mathcal{|B|}} \mathbb{S}(\mathcal{G}(Q^{(i)}), \mathcal{D}(I^{(j)}_{t})) },
    \label{eq:3}
    \end{equation}
\vskip 0.01in
\end{small}%
where $\mathcal{G}(\cdot)$ denotes the feature encoding process of the GM branch.
The overall loss function for the joint fine-tuning stage is as follows:
\begin{equation}
    \mathcal{L}_{Joint} = \mathcal{L}_{D \! I} + \gamma \, \mathcal{L}_{G \! M},
\label{eq:4}
\end{equation}
where $\gamma$ is a trade-off hyper-parameter for the loss functions of the two branches.

\textbf{Training of Compositor.} To facilitate multi-granularity information interaction and encode the multimodal query into a unified feature, we design a lightweight Compositor and train it from scratch: \par
\vspace{-0.8\baselineskip}
\begin{small}
    \begin{equation}
        \mathcal{L}_{C} = - \frac{1}{\mathcal{|B|}} \sum_{i=1}^{\mathcal{|B|}} \log
        \frac{\mathbb{S}(\mathcal{C}(\mathcal{G}(Q^{(i)}), \mathcal{D}(Q^{(i)})), \mathcal{D}(I^{(i)}_{t}))}
        {\sum_{j = 1}^{\mathcal{|B|}} \mathbb{S}(\mathcal{C}(\mathcal{G}(Q^{(i)}), \mathcal{D}(Q^{(i)})), \mathcal{D}(I^{(j)}_{t}))},
    \label{eq:5}
    \end{equation}
\vskip 0.01in
\end{small}%
where $\mathcal{C}(\cdot)$ denotes the feature fusion process of the Compositor. During this training stage, the parameters of both the DI and GM branches are frozen, allowing the Compositor to learn how to dynamically fuse global and detailed features tailored to each multimodal query. By leveraging the complementary strengths of the two branches, the Compositor effectively integrates comprehensive semantic information with fine-grained details, thereby enhancing the overall retrieval performance.

\begin{figure}[t]
\begin{center}
\centerline{\includegraphics[width=1.02\linewidth, trim=0.38cm 0.52cm 0.32cm 0cm, clip]{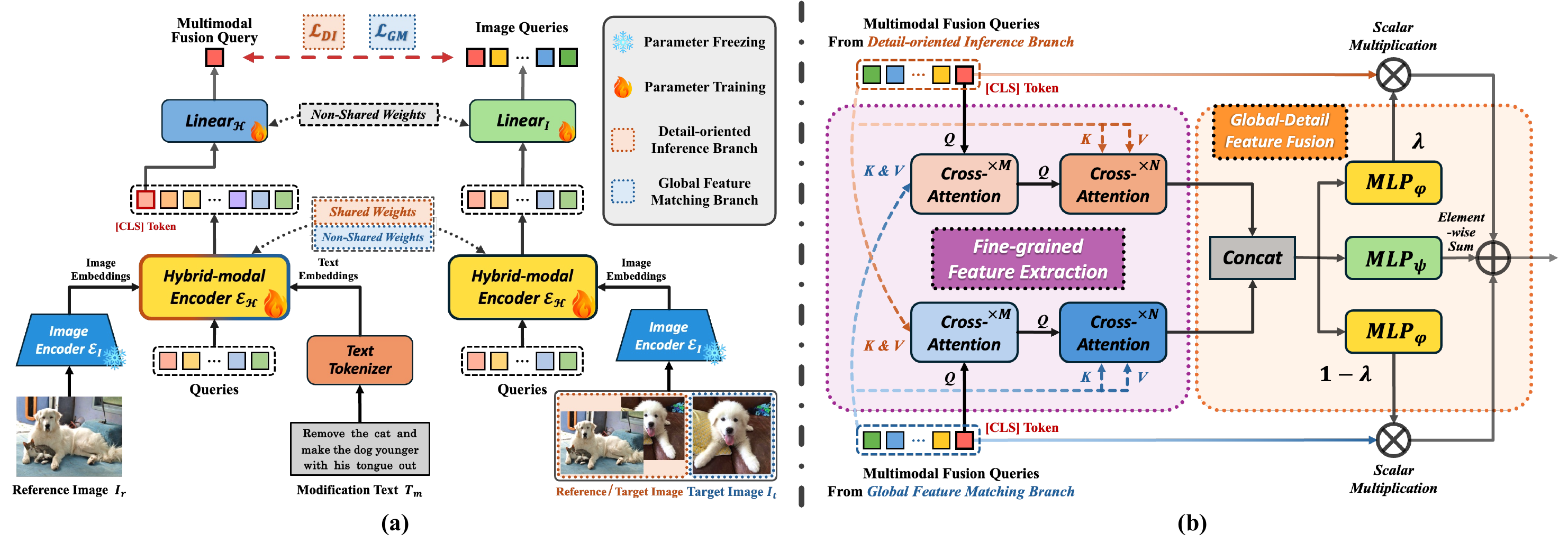}}
\vspace{-0.25\baselineskip}
\caption{Module architecture of our proposed \textbf{\textit{DetailFusion}}. (a) Illustration of the shared structure between the \textit{DI and GM branches}. A hybrid-modal encoder separately encodes the query and image, corresponding to the left and right branches, respectively. The image encoder shares parameters with the DI branch, but non-shared with the GM branch. (b) Illustration of the \textit{Adaptive Feature Compositor}. Features from both branches first interact with fine-grained tokens from both the opposite and the same branch through cross-attention layers within the \textit{Fine-grained Feature Extraction Block}, followed by a convex combination in the \textit{Global-Detail Feature Fusion Block}.}
\label{fig:module_architecture}
\end{center}
\vskip -0.2in
\end{figure}

\subsection{Module Architecture}
\label{subsec:module_architecture}

The DI and GM branches share an identical architecture, differing in their input data and optimization strategies, as shown in Fig.~\ref{fig:module_architecture}(a).
Both branches process multimodal queries through fusion encoding to generate global or detailed features.
Additionally, the DI branch also serves as the image encoder, extracting features directly from images.
The core component is the hybrid-modal encoder $\mathcal{E_H}$, which integrates information from both modalities into a set of learnable query tokens using an attention mechanism. These tokens are then projected into a unified feature space through a linear layer.

For the multimodal query, the feature encoding process of both branches is formulated as follows:
\begin{equation}
    \mathcal{D}(Q) = \mathcal{G}(Q) = \textit{}{Linear}_\mathcal{H} (\mathcal{E_H}(\mathcal{E_I}(I_r), T_m)),
\label{eq:6}
\end{equation}

For the image-only input, the feature encoding process of the DI branch is formulated as follows:
\begin{equation}
    \mathcal{D}(I) = \textit{}{Linear}_\mathcal{I} (\mathcal{E_H}(\mathcal{E_I}(I))),
\label{eq:7}
\end{equation}

where $\textit{Linear}_\mathcal{H}(\cdot)$ denotes the hybrid-modal mapping layer, $\mathcal{E_I}$ denotes the frozen image encoder, $\textit{Linear}_\mathcal{I}(\cdot)$ denotes the image feature mapping layer, and $I$ denotes the single image.

As shown in Fig.~\ref{fig:module_architecture}(b), our designed Adaptive Feature Compositor consists of two primary components: the \textit{Fine-grained Feature Extraction Block} and the \textit{Global-Detail Feature Fusion Block}.
The Feature Extraction block is designed to extract fine-grained information embedded within the query tokens from both branches, which is subsequently injected into the overall representation [CLS] token. This block also facilitates fine-grained interactions between global and detailed features, providing valuable prior knowledge for the subsequent fusion process. For an encoded feature from any branch, the process begins by enabling the [CLS] token to interact with all feature tokens from the opposite branch via cross-attention, allowing it to capture complementary information. The [CLS] token then interacts with all feature tokens from its own branch through additional cross-attention layers, further refining relevant fine-grained details and extracting useful information.

The Feature Fusion block follows the general feature combiner structure~\cite{clip4cir}, aiming to produce a convex combination of the global and detailed features. A notable distinction here is the elimination of projection layers, as both features have already been roughly aligned with the target image features and reside within the same feature space.
The features are concatenated and passed through two separate MLP layers. ${M\!L\!P}_{\!\varphi}$ computes the coefficients for the combination of two features, while ${M\!L\!P}_{\!\psi}$ generates a bridging feature to facilitate a smooth fusion. Finally, the convex combination of the query features and the bridging feature are element-wise summed to form the final representation.

\section{Experiments}
\label{sec:experiments}

\subsection{Experimental Setup}
\label{subsec:experimental_setup}

\textbf{Datasets and Evaluation Metrics.}
Following previous work, we evaluate our method on two widely adopted CIR benchmarks, CIRR~\cite{cirr} and FashionIQ~\cite{fashioniq}.
The CIRR dataset comprises 36,554 triplets derived from 21,552 real-life images sourced from the NLVR2 dataset~\cite{nlvr2}. It also includes a specialized subset designed to evaluate fine-grained discrimination, with each image set comprising six highly similar images to serve as the search space.
The FashionIQ dataset contains 30,134 triplets created from 77,684 web-crawled images in the fashion domain. The dataset is divided into three categories based on fashion product types: Dress, Shirt, and Toptee.
For performance comparison, we report the average recall at rank K (Recall@K). For CIRR, we present Recall@1, 5, 10 and 50, along with $\text{Recall}_\text{subset}\text{@}$1, 2 and 3 respectively reflect the model’s global retrieval and fine-grained detail discrimination capabilities. For FashionIQ, we report Recall@10 and 50 across the three categories to assess the model's generalization ability within specific fashion domains.

\textbf{Implementation Details.}
All the experiments are conducted on a single Tesla V100 GPU with 32GB of memory. The hybrid-modal encoder is initialized from the BLIP-2 pre-trained Q-Former~\cite{blip2}, while the vision encoder is the frozen ViT-G/14 from EVA-CLIP~\cite{eva_clip}. In the first training stage, the epoch is set to 2, the batch size is set to 128, and the best-performing iteration on CIRR and FashionIQ is selected as the initialization for the next stage. In the second stage, the epoch is set to 50, and the batch size is set to 80. In the third stage, the epoch is set to 100 with a batch size of 512. The AdamW optimizer~\cite{adamw} is used throughout training, with betas set to (0.9, 0.98) and a weight decay of 0.05. The learning rate for the first two stages is set to 1e-5, and it is set to 1e-5 and 2e-5 for CIRR and FashionIQ in the third stage, respectively. The temperature hyper-parameters $\tau$ is set to 0.07.

\begin{table}[t]
\vspace{-1.5\baselineskip}
\caption{\textbf{Quantitative comparison with state-of-the-art methods on the FashionIQ validation~set and CIRR test set.} For FashionIQ, ``Avg.'' represents the average results across all evaluation metrics. For CIRR, ``Avg.'' refers to (Recall@5 + $\text{Recall}_\text{subset}\text{@1}$) / 2, indicating the average results across the global and subset settings. The best result is highlighted in \textbf{bold}, while the second best is \underline{underlined}. † indicates the use of an extra training method, Scaling Positives and Negatives (SPN)~\cite{spn}.}
\label{tab:fiq_cirr_results}
\begin{center}
\begin{Huge}
\renewcommand{\arraystretch}{1.102}
\resizebox{\linewidth}{!}{
\begin{tabular}{lc|cc|cc|cc|c|cccc|ccc|c}
\toprule
\multirow{3.5}{*}{\,\,\textbf{Method}} & \multirow{3.5}{*}{\textbf{Publication}} & \multicolumn{7}{c|}{\textbf{FashionIQ}} & \multicolumn{8}{c}{\textbf{CIRR}} \\
\cmidrule(lr){3-9} \cmidrule(lr){10-17}
 & & \multicolumn{2}{c|}{\textbf{Dress}} & \multicolumn{2}{c|}{\textbf{Shirt}} & \multicolumn{2}{c|}{\textbf{Toptee}} & \multirow{2.5}{*}{{\textbf{Avg.}}} & \multicolumn{4}{c|}{\textbf{$\text{Recall@K}$}} & \multicolumn{3}{c|}{\textbf{$\text{Recall}_\text{subset}\text{@K}$}} & \multirow{2.5}{*}{\textbf{Avg.}} \\
\cmidrule(lr){3-4} \cmidrule(lr){5-6} \cmidrule(lr){7-8} \cmidrule(lr){10-13} \cmidrule(lr){14-16}
 & & \textbf{R@10} & \textbf{R@50} & \textbf{R@10} & \textbf{R@50} & \textbf{R@10} & \textbf{R@50} & & \textbf{K=1} & \textbf{K=5} & \textbf{K=10} & \textbf{K=50} & \textbf{K=1} & \textbf{K=2} & \textbf{K=3} & \\
\midrule
\,\,TIRG~\cite{tirg} & CVPR 2019 & 14.87 & 34.66 & 18.26 & 37.89 & 19.08 & 39.62 & 27.40 &
14.61 & 48.37 & 64.08 & 90.03 & 22.67 & 44.97 & 65.14 & 35.52 \\
\,\,ARTEMIS~\cite{artemis} & ICLR 2022 & 27.16 & 52.40 & 21.78 & 43.64 & 29.20 & 53.83 & 38.04 &
16.96 & 46.10 & 61.31 & 87.73 & 39.99 & 62.20 & 75.67 & 43.05 \\
\,\,CLIP4CIR~\cite{clip4cir} & ACM MM 2023 & 33.81 & 59.40 & 39.99 & 60.45 & 41.41 & 65.37 & 50.07 &
38.53 & 69.98 & 81.86 & 95.93 & 68.19 & 85.64 & 94.17 & 69.09 \\
\,\,CompoDiff~\cite{compodiff} & TMLR 2024 & 40.65 & 57.14 & 36.87 & 57.39 & 43.93 & 61.17 & 49.53 &
22.35 & 54.36 & 73.41 & 91.77 & 35.84 & 56.11 & 76.60 & 29.10 \\
\,\,SSN~\cite{ssn} & AAAI 2024 & 34.36 & 60.78 & 38.13 & 61.83 & 44.26 & 69.05 & 51.40 &
43.91 & 77.25 & 86.48 & 97.45 & 71.76 & 88.63 & 95.54 & 74.51 \\
\,\,BLIP4CIR+Bi~\cite{blip4cir+bi} & WACV 2024 & 42.09 & 67.33 & 41.76 & 64.28 & 46.61 & 70.32 & 55.40 &
40.15 & 73.08 & 83.88 & 96.27 & 72.10 & 88.27 & 95.93 & 72.59 \\
\,\,CaLa~\cite{cala} & SIGIR 2024 & 42.38 & 66.08 & 46.76 & 68.16 & 50.93 & 73.42 & 57.96 &
49.11 & 81.21 & 89.59 & 98.00 & 76.27 & 91.04 & 96.46 & 78.74 \\
\,\,TG-CIR~\cite{tgcir} & ACM MM 2023 & 45.22 & 69.66 & 52.60 & 72.52 & 56.14 & 77.10 & 62.21 &
45.25 & 78.29 & 87.16 & 97.30 & 72.84 & 89.25 & 95.13 & 75.57 \\
\,\,CASE~\cite{dataroaming} & AAAI 2024 & 47.44 & 69.36 & 48.48 & 70.23 & 50.18 & 72.24 & 59.74 &
49.35 & 80.02 & 88.75 & 97.47 & 76.48 & 90.37 & 95.71 & 78.25 \\
\,\,CoVR-BLIP~\cite{covr} & AAAI 2024 & 44.55 & 69.03 & 48.43 & 67.42 & 52.60 & 74.31 & 59.39 &
49.69 & 78.60 & 86.77 & 94.31 & 75.01 & 88.12 & 93.16 & 80.81 \\
\,\,Re-ranking~\cite{reranking} & TMLR 2024 & 48.14 & 71.43 & 50.15 & 71.25 & 55.23 & 76.80 & 62.15 &
50.55 & 81.75 & 89.78 & 97.18 & 80.04 & 91.90 & 96.58 & 80.90 \\
\,\,SPRC~\cite{sprc} & ICLR 2024 & 49.18 & 72.43 & 55.64 & 73.89 & 59.35 & 78.58 & 64.85 &
51.96 & 82.12 & 89.74 & 97.69 & 80.65 & 92.31 & 96.60 & 81.39 \\
\,\,SPRC†~\cite{spn} & ACM MM 2024 & 50.57 & \textbf{74.12} & 57.70 & 75.27 & 60.84 & \underline{79.96} & 66.41 &
\underline{55.06} & 83.83 & 90.87 & 98.29 & \underline{81.54} & 92.65 & 97.04 & 82.69 \\
\,\,CIR-LVLM~\cite{cir_lvlm} & AAAI 2025 & 50.42 & 73.57 & \textbf{58.59} & \underline{75.86} & 59.61 & 78.99 & 66.17 &
53.64 & 83.76 & 90.60 & 97.93 & 79.12 & 92.33 & 96.67 & 81.44 \\
\,\,CCIN~\cite{ccin} & CVPR 2025 & 49.38 & 72.58 & 55.93 & 74.14 & 57.93 & 77.56 & 64.59 &
53.41 & 84.05 & 91.17 & 98.00 & - & - & - & - \\
\midrule
\rowcolor{Compositor}
\multicolumn{2}{l|}{\,\,\textbf{\textit{DetailFusion} (Ours)}} & \underline{50.92} & 73.82 & 57.90 & 75.81 & \underline{60.94} & 79.60 & \underline{66.50}  &  54.55 & \underline{84.19} & \underline{91.35} & \underline{98.51} & 81.52 & \underline{93.23} & \underline{97.18} & \underline{82.86} \\
\rowcolor{Compositor}
\multicolumn{2}{l|}{\,\,\textbf{\textit{DetailFusion}† (Ours)}} & \textbf{51.34} & \underline{74.05} & \underline{58.12} & \textbf{75.95} & \textbf{61.22} & \textbf{80.09} & \textbf{66.79}  &  \textbf{55.76} & \textbf{84.77} & \textbf{91.66} & \textbf{98.58} & \textbf{82.22} & \textbf{93.66} & \textbf{97.50} & \textbf{83.50} \\
\bottomrule
\end{tabular}
}
\end{Huge}
\end{center}
\vspace{-1.2\baselineskip}
\end{table}

\subsection{Comparison with State-of-the-arts}
\label{subsec:comparison_with_sotas}

Table~\ref{tab:fiq_cirr_results} presents the performance of our \textbf{\textit{DetailFusion}} compared to state-of-the-art methods on the FashionIQ validation set and the CIRR test set. It can be seen that \textbf{\textit{DetailFusion}} outperforms previous methods across nearly all metrics, with a substantial margin.
From Table~\ref{tab:fiq_cirr_results}, we observe that:
\textbf{(1)} On CIRR, compared to SPRC~\cite{sprc}, which also employs the BLIP-2 Q-Former~\cite{blip2} backbone, our method achieves improvements of \textbf{+2.59}, \textbf{+2.07}, and \textbf{+0.87} in Recall@1, Recall@5, and $\text{Recall}_\text{subset}\text{@}$1, respectively. Compared to CCIN~\cite{ccin}, our method achieves improvements of \textbf{+1.14} in Recall@1. On FashionIQ, our method surpasses both SPRC and CCIN by \textbf{+1.65} and \textbf{+1.91} in average Recall, with particularly notable gains in the Shirt and Toptee categories. These results not only highlight the effectiveness of managing information at different granularities using separate branches, but also demonstrate that our detail enhancement is highly effective in the fashion domain.
\textbf{(2)} Compared to CIR-LVLM~\cite{cir_lvlm}, we achieve a \textbf{+2.40} improvement in $\text{Recall}_\text{subset}\text{@}$1 and an overall improvement of \textbf{+1.42}, indicating the superior ability of VLP models over MLLM in feature encoding, while enabling detail enhancement during retrieval.
\textbf{(3)} We fine-tune our method using a variant of Scaling Positives and Negatives (SPN)~\cite{spn}, which is designed to separately enhance the capability of both branches, thus yielding further performance improvements. Compared to SPRC with SPN method, we achieve an average improvement of \textbf{+0.81}, validating the robustness of our model. Detailed descriptions of SPN method for the three-stage training are provided in \textbf{Appendix~\ref{sec:appendix_spn}}, and more comprehensive comparative results and detailed analyses on both datasets can be found in \textbf{Appendix~\ref{sec:appendix_quantitative_results}}.

\subsection{Ablation Study}
\label{subsec:ablation_study}

We discuss the impact of pre-training data, the effectiveness of the overall framework design, and the necessity of the three-stage training strategy and loss functions of \textbf{\textit{DetailFusion}} in this section. All ablation experiments are conducted on the CIRR validation set, enabling us to clearly assess the contributions of various settings to the model’s global semantic extraction and fine-grained discrimination capabilities through the global and subset retrieval performance, respectively. For experiments that do not involve the Compositor or the third training stage, we sum the similarity~scores of the query with all candidate images from both branches after joint fine-tuning and then sort to obtain the retrieval results.
Complete results with more ablation analysis are provided in \textbf{Appendix~\ref{sec:appendix_experiment_ablation}}.

\begin{table}[t]
\vspace{-1.5\baselineskip}
\caption{\textbf{Ablation Studies on the CIRR validation set.} Results assessing the impact of pre-training with image editing data, the effectiveness of the dual-branch framework design, the three-stage training strategy, the detail-oriented inference loss function, and the Compositor block architecture.}
\label{tab:ablation_studies}
\begin{center}

\begin{minipage}[t]{\linewidth}
    \subfloat[][Ablation of the \textbf{pre-training} and \textbf{dual-branch framework architecture}.]{
    \resizebox{0.48\linewidth}{!}{
        \begin{Huge}
        \begin{tabular}{ccc|cc|cc|c}
        \toprule
        \multirow{2.4}{*}{ \textbf{Method} } & \multirow{2.4}{*}{\textbf{Pre-training}} & & \multicolumn{2}{c|}{\textbf{$\text{Recall@K}$}} & \multicolumn{2}{c|}{\textbf{$\text{Recall}_\text{s}\text{@K}$}} & \multirow{2.4}{*}{\textbf{Avg.}} \\
        \cmidrule{4-7}{} & & & K=1 & K=5 & K=1 & K=2 \\
        \midrule
            SPRC~\cite{sprc} & \textcolor{red}{\XSolidBrush} & & 54.32 & 84.32 & 81.00 & 93.02 & 82.66 \\
            SPRC~\cite{sprc} & \textcolor{teal}{\Checkmark} & & 54.57 & 84.43 & 81.35 & 93.17 & 82.89 \\
            \rowcolor{GM}
            Only \textit{GM} Branch & \textcolor{red}{\XSolidBrush} & & 55.05 & 84.86 & 79.19 & 92.49 & 82.03 \\
            \rowcolor{DI}
            Only \textit{DI} Branch & \textcolor{teal}{\Checkmark} & & 53.69 & 83.89 & 81.73 & 93.67 & 82.81 \\
            \rowcolor{Compositor}
            \textbf{\textit{DetailFusion}} & \textcolor{teal}{\Checkmark} & & \textbf{56.83} & \textbf{85.75} & \textbf{82.28} & \textbf{94.07} & \textbf{84.01} \\
        \bottomrule
        \end{tabular}
        \end{Huge}
        }
    \label{tab:ablation_data_framework}
    }
    \hfill
    \subfloat[][Ablation of the \textbf{training strategies} for the dual-branch framework.]{
    \renewcommand{\arraystretch}{1.1}
    \resizebox{0.48\linewidth}{!}{
        \begin{large}
        \begin{tabular}{c|cc|cc|c}
        \toprule
        \multirow{2.5}{*}{\textbf{Training strategy}} & \multicolumn{2}{c|}{\textbf{$\text{Recall@K}$}} & \multicolumn{2}{c|}{\textbf{$\text{Recall}_\text{s}\text{@K}$}} & \multirow{2.5}{*}{\textbf{Avg.}} \\
        \cmidrule{2-5}{} & K=1 & K=5 & K=1 & K=2 \\
        \midrule
            Only Stage 2 & 55.03 & 84.88 & 80.72 & 93.28 &	82.80 \\
            Hybrid of Stage 1\&2 & 54.84 & 84.81 & 81.25 & 93.45 &	83.03 \\
            \rowcolor{gray!15}
            Stage 1 \& Stage 2 &
            55.92 & 85.29 & 81.63 & 93.71 &	83.46 \\
            \rowcolor{Compositor}
            \textbf{\textit{DetailFusion} (Full)} & \textbf{56.83} & \textbf{85.75} & \textbf{82.28} & \textbf{94.07} & \textbf{84.01} \\
        \bottomrule
        \end{tabular}
        \end{large}
        }
    \label{tab:ablation_training_stage}
    }
\end{minipage}
\smallskip
\begin{minipage}[t]{\linewidth}
    \subfloat[][Ablation of the \textbf{loss functions} for the \textit{Detail-oriented Inference branch} during training.]{
    \renewcommand{\arraystretch}{1.1}
    \resizebox{0.48\linewidth}{!}{
        \begin{large}
        \begin{tabular}{cc|cc|cc|c}
        \toprule
        \multirow{2.5}{*}{\textbf{Stage 1}} & \multirow{2.5}{*}{\textbf{Stage 2}} & \multicolumn{2}{c|}{\textbf{$\text{Recall@K}$}} & \multicolumn{2}{c|}{\textbf{$\text{Recall}_\text{s}\text{@K}$}} & \multirow{2.5}{*}{\textbf{Avg.}} \\
        \cmidrule{3-6}{} & & K=1 & K=5 & K=1 & K=2 \\
        \midrule
            \textcolor{red}{$\mathcal{L}_{G \! M}$} & \textcolor{red}{$\mathcal{L}_{G \! M}$} & 55.13 & 84.76 & 79.50 & 92.54 &	82.13 \\
            \textcolor{teal}{$\mathcal{L}_{D \! I}$} & \textcolor{red}{$\mathcal{L}_{G \! M}$} & 55.47 & 85.05 & 80.24 & 92.78 &	82.65 \\
            \textcolor{red}{$\mathcal{L}_{G \! M}$} & \textcolor{teal}{$\mathcal{L}_{D \! I}$} & 55.51 & 85.24 & 81.20 & 93.47 &	83.22 \\
            \rowcolor{gray!15}
            \textcolor{teal}{$\mathcal{L}_{D \! I}$} & \textcolor{teal}{$\mathcal{L}_{D \! I}$} & \textbf{55.92} & \textbf{85.29} & \textbf{81.63} & \textbf{93.71} & \textbf{83.46} \\
        \bottomrule
        \end{tabular}
        \end{large}
        }
    \label{tab:ablation_loss_function}
    }
    \hfill
    \subfloat[][Ablation of the \textbf{block architectures} for the \textit{Adaptive Feature Compositor}.]{
    \resizebox{0.48\linewidth}{!}{
        \begin{Huge}
        \begin{tabular}{cc|cc|cc|c}
        \toprule
        \multirow{2.4}{*}{ \textbf{Extraction} } & \multirow{2.4}{*}{\textbf{Fusion}} & \multicolumn{2}{c|}{\textbf{$\text{Recall@K}$}} & \multicolumn{2}{c|}{\textbf{$\text{Recall}_\text{s}\text{@K}$}} & \multirow{2.4}{*}{\textbf{Avg.}} \\
        \cmidrule{3-6}{} & & K=1 & K=5 & K=1 & K=2 \\
        \midrule
            \rowcolor{gray!15}
            - & \textit{Average} & 55.92 & 85.29 & 81.63 & 93.71 & 83.46 \\
            \textit{Concat} & \multirow{2}{*}{\textcolor{Dandelion}{\textbf{${M\!L\!P}_{\!\varphi}$}} \!\!\! \textbf{+} \!\! \textcolor{ForestGreen}{\textbf{${M\!L\!P}_{\!\psi}$}}} & 56.52 & 85.42 & 81.87 & 93.77 & 83.64 \\
            \textit{Projection} & & 56.71 & 85.67 & 82.01 & 93.85 &	83.84 \\
            \multirow{2}{*}{\textbf{\textit{Attention}}} & \textcolor{Dandelion}{\textbf{${M\!L\!P}_{\!\varphi}$}} & 55.85 & 85.19 & 81.75 & 93.76 &	83.47 \\
             & \textcolor{ForestGreen}{\textbf{${M\!L\!P}_{\!\psi}$}} &
            55.56 & 85.17 & 81.39 & 93.33 & 83.28 \\
            \rowcolor{Compositor}
            \textbf{\textit{Attention}} & \textcolor{Dandelion}{\textbf{${M\!L\!P}_{\!\varphi}$}} \!\!\! \textbf{+} \!\! \textcolor{ForestGreen}{\textbf{${M\!L\!P}_{\!\psi}$}} & \textbf{56.83} & \textbf{85.75} & \textbf{82.28} & \textbf{94.07} & \textbf{84.01} \\
        \bottomrule
        \end{tabular}
        \end{Huge}
        }
    \label{tab:ablation_block_architecture}
    }
\end{minipage}
\end{center}
\vspace{-2.5\baselineskip}
\end{table}

\textbf{Fairness of Using Image Editing Data for Pre-Training.}
To validate the effectiveness of our~method in subsequent ablation experiments, and to ensure that performance gains are not merely due to the increased scale of supervised data, we pre-trained SPRC~\cite{sprc} under the same dataset and training strategy, and then evaluated on the CIRR validation set for a fair comparison, as shown in Table~\ref{tab:ablation_studies}(a).  Comparing the first and second rows shows that even the reference images are used as hard negatives during pre-training, the performance improvements of SPRC remain marginal. These results indicate
\begin{wrapfigure}{r}{0.48\linewidth}
\vskip 0.2in
\vspace{-1.6\baselineskip}
\begin{center}
\centerline{\includegraphics[width=\linewidth, trim=0.25cm 0.35cm 0.38cm 0.35cm, clip]{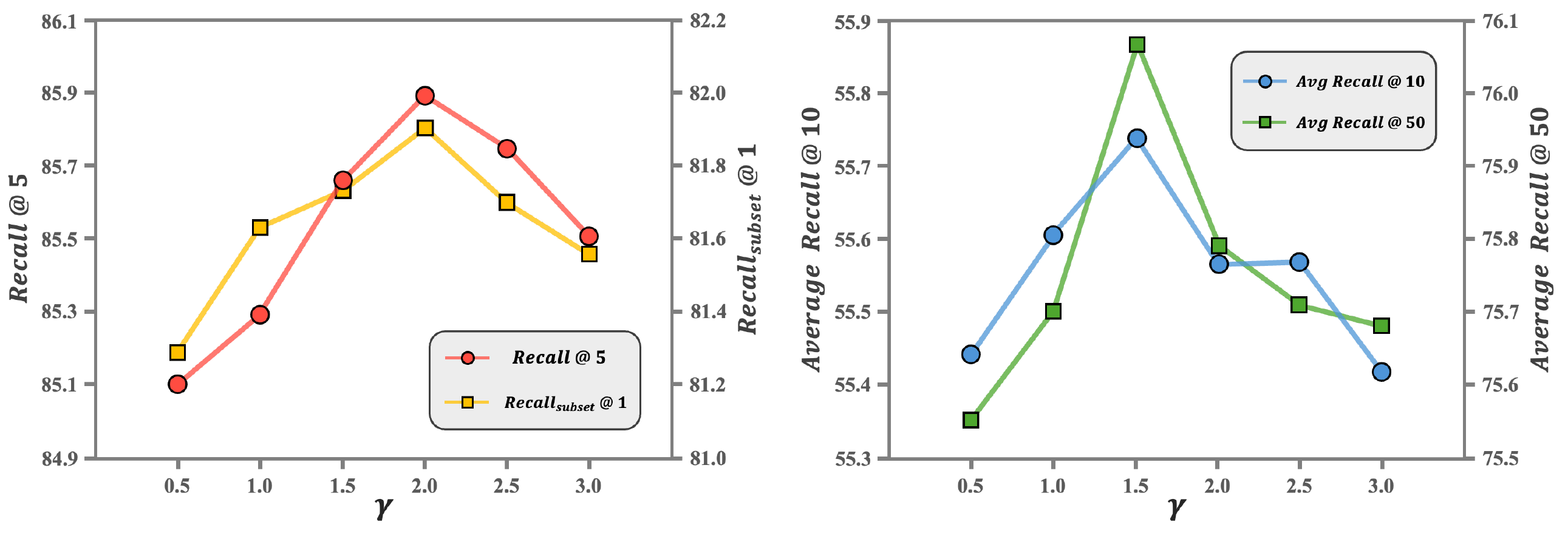}}
\vspace{-0.25\baselineskip}
\caption{Analysis of different values of \textbf{trade-off hyper-parameter} $\mathbf{\gamma}$ during joint fine-tuning.}
\label{fig:ablation_gamma}
\end{center}
\vspace{-2\baselineskip}
\end{wrapfigure}
that the image editing dataset, due to its distributional deviation, is not suitable for direct CIR pre-training. However, by incorporating this dataset into our method, we effectively stimulate the model’s capacity for detail perception and processing, yielding significant performance gains.

\textbf{Complementarity of Dual-Branch Function.}
According to Table~\ref{tab:ablation_studies}(a), the GM branch exhibits higher overall retrieval accuracy, while the DI branch performs superior subset retrieval performance. Fusing features from both branches with the Compositor yields improved results across the entire benchmark. These results indicate that the two branches possess complementary functionalities: the GM branch excels at capturing global semantic, while the DI branch specializes in fine-grained detail discrimination. Based on these, \textbf{\textit{DetailFusion}} effectively integrates the strengths of both branches through the \textit{Adaptive Feature Compositor}.

\textbf{Effectiveness and Necessity of the Three-Stage Training Strategy.}
According to Table~\ref{tab:ablation_studies}(b), pre-training the DI branch in the first stage followed by joint fine-tuning of both branches achieves improvements of \textbf{+0.89} and \textbf{+0.91} in Recall@1 and $\text{Recall}_\text{subset}\text{@}$1, compared to directly training both branches, which validates the effectiveness of the first stage.
On the other hand, mixing the IPr2Pr and CIRR datasets for directly second stage training causes a severe decrease in overall performance due to the data distributional deviation, confirming the necessity of the first stage.
Furthermore, implementing fine-grained feature fusion through the Compositor produces further improvements of \textbf{+0.91} and \textbf{+0.65} in Recall@1 and $\text{Recall}_\text{subset}\text{@}$1, validating the effectiveness of the third stage.

\textbf{Discussion on \textit{DI} Branch Loss Function.}
As we consistently optimize the DI branch using $\mathcal{L}_{D \! I}$, we compare the results of training the DI branch using the batch-based contrastive loss $\mathcal{L}_{G \! M}$ during the first two stages. As shown in Table~\ref{tab:ablation_studies}(c), applying $\mathcal{L}_{G \! M}$ to optimize the DI branch at any stage leads to a decline in retrieval accuracy, particularly on the subset metrics. These results demonstrate that treating the reference images as hard negatives during contrastive learning is crucial for enhancing and preserving the DI branch's detail capabilities. Applying the same optimization strategy to both branches causes their functionalities to converge, thereby weakening the complementary advantages.

\textbf{Discussion on Block Architectures in \textit{Adaptive Feature Compositor}.}
Our Compositor employs an attention mechanism within the Feature Extraction Block to facilitate fine-grained information injection. We also experimented with concatenation and projection for combining the two features, as referenced in the general combiner structure~\cite{clip4cir}. As shown in Table~\ref{tab:ablation_studies}(d), our attention mechanism achieves the best performance, since the features before fused are already reside in the same feature space, eliminating the need for additional alignment using projection layers.
Furthermore, within the Feature Fusion Block, ${M\!L\!P}_{\!\varphi}$ produces the fusion ratio, and ${M\!L\!P}_{\!\psi}$ generates the bridging feature~for a convex combination. Removing either MLP block causes a significant performance drop. With only ${M\!L\!P}_{\!\varphi}$, the fine-grained information cannot be integrated into the final representation. Conversely, with only ${M\!L\!P}_{\!\psi}$, the network fails to converge during training without relying on the original features.

\textbf{Analysis of trade-off in Joint Fine-tuning Loss.}
In the second training stage, to ensure that both branches are optimized cooperatively at nearly the same rate, it is necessary to adjust the weight of the loss function using the trade-off hyper-parameter $\gamma$. We record the retrieval performance for different values of $\gamma$ on two distinct datasets, as shown in Fig.~\ref{fig:ablation_gamma}. As observed, with the increase of $\gamma$, the retrieval accuracy initially rises gradually, reaches a peak, and then begins to decline. Therefore, based on our experiments, we set $\gamma\!=\!2.0$ for the CIRR dataset and $\gamma\!=\!1.5$ for the FashionIQ dataset to achieve the best retrieval performance. The complete results are provided in Table~\ref{tab:ablation_trade_off_cirr} and Table~\ref{tab:ablation_trade_off_fiq}.

\textbf{Discussion on Computational Complexity and Inference Time.}
The pre-training in Stage 1~requires just over one epoch, taking nearly 9.5 hours on our devices. The pretrained weights are general and can be applied for fine-tuning on any CIR dataset in subsequent stages. Stage 2 requires nearly 14.2 hours on the CIRR dataset and nearly 10.7 hours on the FashionIQ dataset. The number of trainable parameters in Stage 3 is only 1\% of those in Stage 2. By pre-computing and storing the Compositor’s input embeddings, the training time for Stage 3 can be reduced to 0.5 hours. Therefore, the overall computational and time overhead during training for \textbf{\textit{DetailFusion}} on CIR datasets is only 10\%-20\% higher than the single-stage methods.
During inference, the dual-branch computation is parallelized, yielding inference times comparable to existing methods: \textbf{\textit{DetailFusion}}’s inference time is approximately 1.5 times that of a single Q-Former~\cite{blip2}, and only 12\% slower than the SPRC~\cite{sprc}.
Quantitative analysis of computational complexity and inference time are provided in \textbf{Appendix~\ref{subsec:appendix_computational_complexity}}.

\begin{figure}[t]
\vspace{-0.5\baselineskip}
\begin{center}
\centerline{\includegraphics[width=1.02\linewidth, trim=0.32cm 0.45cm 0.32cm 0.3cm, clip]{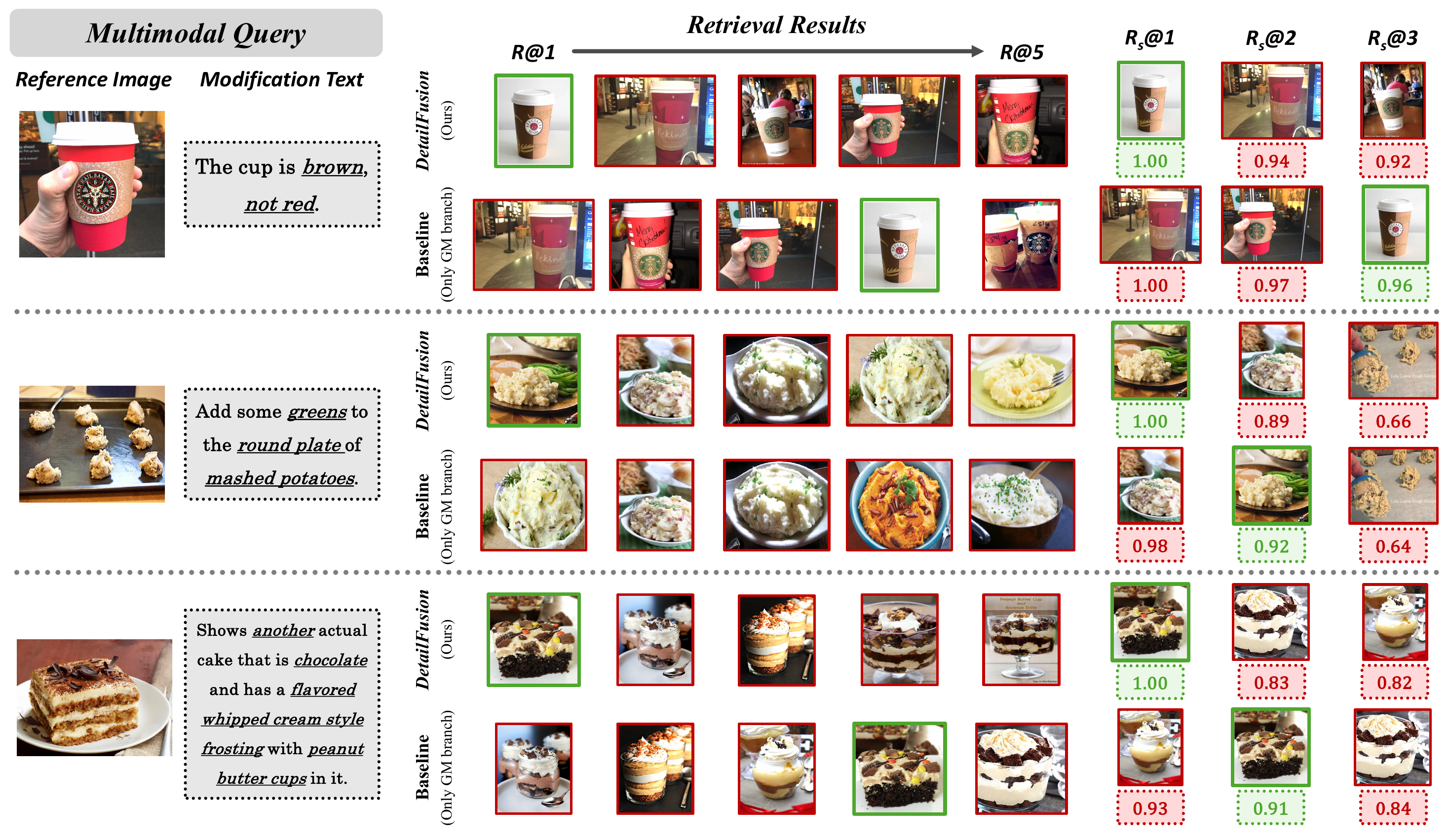}}
\caption{\textbf{Qualitative comparison of our method and the Baseline on the CIRR validation set.} Images are arranged in descending order from left to right based on similarity to the multimodal~query. The \textbf{\textcolor[HTML]{70ad47}{green boxes}} highlight the target image, while all non-target images are marked with \textbf{\textcolor[HTML]{aa0d06}{red boxes}}. In the subset retrieval results, we report the \textit{relative similarity score}, which defined as the normalized similarity between each retrieved image and the multimodal query, relative to all images in the gallery.}
\vspace{-0.5\baselineskip}
\label{fig:qualitative_comparison_baseline}
\end{center}
\vskip -0.2in
\end{figure}

\subsection{Case Study}
\label{subsec:case_study}

Referring to the retrieval results in Fig.~\ref{fig:qualitative_comparison_baseline}, it is evident that our method, with its superior detail capabilities, is better equipped to handle subtle image alterations and complex modification requirements in CIR. According to the relative similarity scores reported in the subset retrieval results, our method excels at differentiating highly semantically similar images, as demonstrated by the clear margin between the correctly retrieved target image and the most similar image. In contrast, the baseline model struggles to retrieve the target image among such confusing images, as these images exhibit closely relative similarity scores. More qualitative comparison results are provided in \textbf{Appendix~\ref{sec:appendix_qualitative_comparison}}.

\section{Conclusion}
\label{sec:conclusion}

In this work, we proposed \textbf{\textit{DetailFusion}}, an innovative dual-branch framework that enhances the capacity to perceive and process fine-grained details within the CIR task. By proposing a detail-oriented optimization strategy, we ingeniously leverage an image editing dataset to develop a branch dedicated to details. By proposing the Feature Compositor, we established coordination between global and detailed information during retrieval. Furthermore, we designed a three-stage training strategy to progressively and optimally refine each component of the framework.
Extensive experiments demonstrate that \textbf{\textit{DetailFusion}} consistently outperforms existing CIR methods.
To the best of our knowledge, we are the first to explore and validate the pivotal role of detail enhancement in CIR.


{
\small
\nocite{conditioned, promptclip, mmt, nsfse, css, pseudo, vqa4cir, finecir, textual_inversion, keds, ldre, seize, cirevl, survey}
\bibliography{DetailFusion_arxiv}
\bibliographystyle{unsrt}
}

\newpage
\appendix

\section{Scaling Positives and Negatives for the Three-Stage Training Strategy}
\label{sec:appendix_spn}

\textbf{Scaling Positives and Negatives (SPN)}~\cite{spn} addresses the limitation of having insufficient positive and negative samples in batch-based contrastive learning, which is constrained by limited batch~size. SPN enhances contrastive learning by providing adequate positive and negative samples, thereby improving the representation quality of the query features. For a model that already trained on CIR datasets, SPN is implemented by freezing the Image Encoder and encoding all images in the candidate set. The model is then fine-tuned again, allowing each query to encounter with all negative samples, effectively simulating an expanded training batch. Additionally, introducing unseen positive samples during the fine-tuning process mitigates overfitting, leading to better generalization performance.

Directly applying the SPN method to our proposed \textbf{\textit{DetailFusion}} framework requires performing the aforementioned fine-tuning steps after both the second and third training stages. However, we observed that this naive approach does not improve performance, which can be attributed to two main reasons: (1) During training, our DI branch treats the reference images as negative samples. This operation partially overlaps with the role of Scaling Negatives, which also aims to increase the diversity of negative samples. (2) SPN treats all non-target images in the candidate set as negatives. The large number of such negative samples dilutes the presence of hard negatives during DI branch training, thereby undermining the effectiveness of pre-training designed to enhance the DI branch’s fine-grained detail perception and processing capabilities.
To address these issues, we modified the SPN method to separately enhance the capabilities of both branches. In this section, we provide a detailed description and analysis of the adapted SPN method tailored for our \textbf{\textit{DetailFusion}} framework.

Since the first training stage involves only the pre-training of the DI branch, the SPN method is not applied at this stage. In the second training stage, our objective is to leverage the variants of the SPN method to independently enhance the global semantic extraction capabilities of the GM branch and the detail perception and processing capabilities of the DI branch. Benefiting from the subset partitioning in the CIRR dataset, each image has five corresponding semantically highly similar images. Utilizing these highly similar images as additional hard negatives to train the DI branch can further improve its detail discrimination ability. We define $[I]_{g}=\{I_{g_{1}}, I_{g_{2}}, ..., I_{g_{5}}\}$ as the set of five highly similar images within the same subset as image $I$. At the beginning of the second training stage, we first apply the \textbf{Scaling Group Negatives (SGN)} method exclusively to the DI branch. The variants of the detail-oriented inference loss function is formulated as follows: \par
\vspace{-0.8\baselineskip}
\begin{small}
    \begin{equation}
        \mathcal{L}_{D \! I}^{\textbf{\,SGN}} = -\frac{1}{\mathcal{|B|}} \sum_{i=1}^{\mathcal{|B|}} \log
        \frac{\mathbb{S}(\mathcal{D}(Q^{(i)}), \mathcal{D}(I^{(i)}_{t}))}
        { \sum_{j=1}^{\mathcal{|B|}}  \mathbb{S}(\mathcal{D}(Q^{(i)}), \mathcal{D}(I^{(j)}_{t})
        \bigoplus \mathcal{D}(I^{(j)}_{r}) \bigoplus \mathcal{D}([I^{(i)}_{t}]_{g}) )},
    \label{eq:8}
    \end{equation}
\vskip 0.01in
\end{small}%
For the GM branch during the joint fine-tuning process, we aim to enable it to acquire a foundational capability through batch-based contrastive learning. Therefore, we preserve its optimization strategy as illustrated in Equation~\ref{eq:3}.
The overall loss function at the second training stage is as follows:
\begin{equation}
    \mathcal{L}_{Joint}^{\textbf{\,SGN}} = \mathcal{L}_{D \! I}^{\textbf{\,SGN}} + \gamma \, \mathcal{L}_{G \! M},
\label{eq:9}
\end{equation}
Subsequently, we freeze the parameters of the DI branch and the Image Encoder, applying SPN for the GM branch. The variants of the global feature matching loss function is formulated as follows: \par
\vspace{-0.8\baselineskip}
\begin{small}
    \begin{equation}
        \mathcal{L}_{G \! M}^{\textbf{\,SPN}} = -\frac{1}{\mathcal{|B|}} \sum_{i=1}^{\mathcal{|B|}} \log
        \frac{\mathbb{S}(\mathcal{G}(Q^{(i)}), \mathcal{D}(I^{(i)}_{t}))}
        { \sum_{I^{(j)}_{t} \! \in \mathbf{G}} \mathbb{S}(\mathcal{G}(Q^{(i)}), \mathcal{D}(I^{(j)}_{t})) },
    \label{eq:10}
    \end{equation}
\vskip 0.01in
\end{small}%
Through the aforementioned training steps, we effectively enhance the individual capabilities of each branch before integrating the Compositor. By incorporating the SPN method during the subsequent training of the Compositor, we are able to activate and fully leverage the strengths of both branches.

In the third training stage, we first freeze the parameters of both branches and the Image Encoder, then train the Compositor from scratch using the loss function $\mathcal{L}_{C}$ as defined in Equation~\ref{eq:5}. After the Compositor acquires fundamental feature fusion capabilities and achieves approximate alignment with the target image features, we finally apply the SPN method to achieve fine-grained alignment. The variants of the Compositor training loss function is formulated as follows: \par
\vspace{-0.8\baselineskip}
\begin{small}
    \begin{equation}
        \mathcal{L}_{C}^{\textbf{\,SPN}} = - \frac{1}{\mathcal{|B|}} \sum_{i=1}^{\mathcal{|B|}} \log
        \frac{\mathbb{S}(\mathcal{C}(\mathcal{G}(Q^{(i)}), \mathcal{D}(Q^{(i)})), \mathcal{D}(I^{(i)}_{t}))}
        { \sum_{I^{(j)}_{t} \! \in \mathbf{G}} \mathbb{S}(\mathcal{C}(\mathcal{G}(Q^{(i)}), \mathcal{D}(Q^{(i)})), \mathcal{D}(I^{(j)}_{t})) },
    \label{eq:11}
    \end{equation}
\end{small}%

\section{Additional Implementation Details and Quantitative Comparison Results}
\label{sec:appendix_quantitative_results}

In this section, we first present the implementation details regarding the use of the SPN method for the three-stage training strategy in Subsection~\ref{subsec:appendix_implementation_details}. Subsequently, we report comprehensive quantitative comparison results on the CIRR test set and the FashionIQ validation set in Subsection~\ref{subsec:appendix_cirr} and Subsection~\ref{subsec:appendix_fiq}, respectively. This enables to highlight the superiority of our proposed method.



\subsection{Implementation Details for the SPN Method}
\label{subsec:appendix_implementation_details}

As described in Section~\textbf{\ref{sec:appendix_spn}}, during the second training stage, we first implement SGN on the DI branch within the joint fine-tuning process. The number of epochs is set to 10, the batch size is set to 50, and the learning rate is set to 1e-5. Subsequently, we freeze the parameters of the DI branch and the Image Encoder, and apply the SPN method to the GM branch, with the number of epochs set to 20, the batch size set to 8, and the learning rate set to 5e-7. In the third training stage, we initially train the Compositor, followed by applying the SPN method to the Compositor. During the SPN for this stage, the number of epochs is set to 5, the batch size is set to 32, and the learning rate is set to 1e-5.

\subsection{Comprehensive Quantitative Comparison on the CIRR dataset}
\label{subsec:appendix_cirr}

\begin{table}[h!]
\vspace{-\baselineskip}
\caption{\textbf{Quantitative comparison with the state-of-the-art methods on the CIRR test set.} ``Avg.'' refers to (Recall@5 + $\text{Recall}_\text{subset}\text{@1}$) / 2, indicating the average results across the global and subset settings. The best result is highlighted in \textbf{bold}, while the second best is \underline{underlined}. * indicates that the method deploys an extra re-ranking strategy. † indicates the use of an extra training method~\cite{spn}.}
\label{tab:cirr_results_full}
\begin{center}
\begin{footnotesize}
\resizebox{\linewidth}{!}{
\begin{tabular}{lccc|cccc|ccc|c}
\toprule
\multirow{2.6}{*}{\textbf{Method}} & \multirow{2.6}{*}{\textbf{Backbone}} & \multirow{2.6}{*}{\textbf{Year}} & & \multicolumn{4}{c|}{\textbf{$\text{Recall@K}$}} & \multicolumn{3}{c|}{\textbf{$\text{Recall}_\text{subset}\text{@K}$}} & \multirow{2.6}{*}{\textbf{Avg.}} \\
\cmidrule(lr){5-8} \cmidrule(lr){9-11}
 & & & & \textbf{K=1} & \textbf{K=5} & \textbf{K=10} & \textbf{K=50} & \textbf{K=1} & \textbf{K=2} & \textbf{K=3} & \\
\midrule
TIRG~\cite{tirg} & w/o VLP & 2019 & & 14.61 & 48.37 & 64.08 & 90.03 & 22.67 & 44.97 & 65.14 & 35.52 \\
MAAF~\cite{maaf} & w/o VLP & 2020 & & 10.31 & 33.03 & 48.30 & 80.06 & 21.05 & 41.81 & 61.60 & 27.04 \\
CIRPLANT~\cite{cirr} & w/o VLP & 2021 & & 19.55 & 52.55 & 68.39 & 92.38 & 39.20 & 63.03 & 79.49 & 45.88 \\
ARTEMIS~\cite{artemis} & w/o VLP & 2022 & & 16.96 & 46.10 & 61.31 & 87.73 & 39.99 & 62.20 & 75.67 & 43.05 \\
CLIP4CIR~\cite{clip4cir} & CLIP & 2023 & & 38.53 & 69.98 & 81.86 & 95.93 & 68.19 & 85.64 & 94.17 & 69.09 \\
TG-CIR~\cite{tgcir} & CLIP & 2023 & & 45.25 & 78.29 & 87.16 & 97.30 & 72.84 & 89.25 & 95.13 & 75.57 \\
CompoDiff~\cite{compodiff} & CLIP & 2024 & & 22.35 & 54.36 & 73.41 & 91.77 & 35.84 & 56.11 & 76.60 & 29.10 \\
SSN~\cite{ssn} & CLIP & 2024 & & 43.91 & 77.25 & 86.48 & 97.45 & 71.76 & 88.63 & 95.54 & 74.51 \\
BLIP4CIR+Bi~\cite{blip4cir+bi} & BLIP & 2024 & & 40.15 & 73.08 & 83.88 & 96.27 & 72.10 & 88.27 & 95.93 & 72.59 \\
CASE~\cite{dataroaming} & BLIP & 2024 & & 49.35 & 80.02 & 88.75 & 97.47 & 76.48 & 90.37 & 95.71 & 78.25 \\
CoVR-BLIP~\cite{covr} & BLIP & 2024 & & 49.69 & 78.60 & 86.77 & 94.31 & 75.01 & 88.12 & 93.16 & 80.81 \\
Re-ranking*~\cite{reranking} & BLIP & 2024 & & 50.55 & 81.75 & 89.78 & 97.18 & 80.04 & 91.90 & 96.58 & 80.90 \\
CaLa~\cite{cala} & BLIP-2 & 2024 & & 49.11 & 81.21 & 89.59 & 98.00 & 76.27 & 91.04 & 96.46 & 78.74 \\
SPRC~\cite{sprc} & BLIP-2 & 2024 & & 51.96 & 82.12 & 89.74 & 97.69 & 80.65 & 92.31 & 96.60 & 81.39 \\
SPRC${^2}$*~\cite{sprc} & BLIP-2 & 2024 & & 54.15 & 83.01 & 90.39 & 98.17 & \textbf{82.31} & 92.68 & 96.87 & 82.66 \\
SPRC†~\cite{spn} & BLIP-2 & 2024 & & \underline{55.06} & 83.83 & 90.87 & 98.29 & 81.54 & 92.65 & 97.04 & 82.69 \\
CCIN~\cite{ccin} & BLIP-2 & 2025 & & 53.41 & 84.05 & 91.17 & 98.00 & - & - & - & - \\
CIR-LVLM~\cite{cir_lvlm} & MLLM & 2025 & & 53.64 & 83.76 & 90.60 & 97.93 & 79.12 & 92.33 & 96.67 & 81.44 \\
\midrule
\rowcolor{Compositor}
\textbf{\textit{DetailFusion} (Ours)} & BLIP-2 & & & 54.55 & \underline{84.19} & \underline{91.35} & \underline{98.51} & 81.52 & \underline{93.23} & \underline{97.18} & \underline{82.86} \\
\rowcolor{Compositor}
\textbf{\textit{DetailFusion}† (Ours)} & BLIP-2 & & & \textbf{55.76} & \textbf{84.77} & \textbf{91.66} & \textbf{98.58} & \underline{82.22} & \textbf{93.66} & \textbf{97.50} & \textbf{83.50} \\
\bottomrule
\end{tabular}
}
\end{footnotesize}
\end{center}
\end{table}

Table~\ref{tab:cirr_results_full} presents the complete results of our method compared to state-of-the-art approaches on~the CIRR test set. Compared to SPRC~\cite{sprc}, our method achieves improvements~of \textbf{+2.59}, \textbf{+2.07}, and \textbf{+0.87} in Recall@1, Recall@5, and $\text{Recall}_\text{subset}\text{@}$1, respectively. Compared to CCIN~\cite{ccin}, our method achieves an improvement of \textbf{+1.14} in Recall@1. Our method slightly outperforms SPRC${^2}$, which adopts complex cross-modal interactions for the re-ranking~\cite{reranking} strategy. Additionally, compared to CIR-LVLM~\cite{cir_lvlm}, we achieve a \textbf{+2.40} improvement in $\text{Recall}_\text{subset}\text{@}$1 and an overall improvement of \textbf{+1.42}. These results confirm that our proposed \textbf{\textit{DetailFusion}} is currently state-of-the-art CIR method.

\subsection{Comprehensive Quantitative Comparison on the FashionIQ dataset}
\label{subsec:appendix_fiq}

\begin{table}[t]
\vspace{-0.5\baselineskip}
\caption{\textbf{Quantitative comparison with state-of-the-art methods on the FashionIQ validation set.} ``Avg.'' refers to (Recall@10 + Recall@50) / 2, indicating the average results across all the evaluation metrics. † indicates the use of an extra training method~\cite{spn}.}
\label{tab:fiq_results_full}
\begin{center}
\begin{large}
\resizebox{\linewidth}{!}{
\begin{tabular}{lccc|cc|cc|cc|cc|c}
\toprule
\multirow{2.6}{*}{\,\textbf{Method}} & \multirow{2.6}{*}{\textbf{Backbone}} & \multirow{2.6}{*}{\textbf{Year}} & & \multicolumn{2}{c|}{\textbf{Dress}} & \multicolumn{2}{c|}{\textbf{Shirt}} & \multicolumn{2}{c|}{\textbf{Toptee}} & \multicolumn{2}{c|}{\textbf{Average}} & \multirow{2.6}{*}{\textbf{Avg.}} \\
\cmidrule(lr){5-6} \cmidrule(lr){7-8} \cmidrule(lr){9-10} \cmidrule(lr){11-12}
 & & & & \textbf{R@10} & \textbf{R@50} & \textbf{R@10} & \textbf{R@50} & \textbf{R@10} & \textbf{R@50} & \textbf{R@10} & \textbf{R@50} & \\
\midrule
\,TIRG~\cite{tirg} & w/o VLP & 2019 & & 14.87 & 34.66 & 18.26 & 37.89 & 19.08 & 39.62 & 17.40 & 37.39 & 27.40 \\
\,MAAF~\cite{maaf} & w/o VLP & 2020 & & 23.80 & 48.60 & 21.30 & 44.20 & 27.90 & 53.60 & 24.30 & 48.80 & 36.60 \\
\,CIRPLANT~\cite{cirr} & w/o VLP & 2021 & & 17.45 & 40.41 & 17.53 & 38.81 & 61.64 & 45.38 & 18.87 & 41.53 & 30.20 \\
\,CosMo~\cite{cosmo} & w/o VLP & 2021 & & 25.64 & 50.30 & 24.90 & 49.18 & 29.21 & 57.46 & 26.58 & 52.31 & 39.45 \\
\,ARTEMIS~\cite{artemis} & w/o VLP & 2022 & & 27.16 & 52.40 & 21.78 & 43.64 & 29.20 & 53.83 & 26.05 & 50.29 & 38.04 \\
\,CLIP4CIR~\cite{clip4cir} & CLIP & 2023 & & 33.81 & 59.40 & 39.99 & 60.45 & 41.41 & 65.37 & 38.40 & 61.74 & 50.07 \\
\,FAME-VIL~\cite{famevil} & CLIP & 2023 & & 42.19 & 67.38 & 47.64 & 68.79 & 50.69 & 73.07 & 46.84 & 69.75 & 58.29 \\
\,TG-CIR~\cite{tgcir} & CLIP & 2023 & & 45.22 & 69.66 & 52.60 & 72.52 & 56.14 & 77.10 & 51.32 & 73.09 & 62.21 \\
\,CompoDiff~\cite{compodiff} & CLIP & 2024 & & 40.65 & 57.14 & 36.87 & 57.39 & 43.93 & 61.17 & 40.48 & 58.57 & 49.53 \\
\,MUR~\cite{mur} & CLIP & 2024 & & 32.61 & 61.34 & 33.23 & 62.55 & 41.40 & 72.51 & 35.75 & 65.47 & 50.61 \\
\,SSN~\cite{ssn} & CLIP & 2024 & & 34.36 & 60.78 & 38.13 & 61.83 & 44.26 & 69.05 & 38.92 & 63.89 & 51.40 \\
\,BLIP4CIR+Bi~\cite{blip4cir+bi} & BLIP & 2024 & & 42.09 & 67.33 & 41.76 & 64.28 & 46.61 & 70.32 & 43.49 & 67.31 & 55.40 \\
\,CoVR-BLIP~\cite{covr} & BLIP & 2024 & & 44.55 & 69.03 & 48.43 & 67.42 & 52.60 & 74.31 & 48.53 & 70.25 & 59.39 \\
\,CASE~\cite{dataroaming} & BLIP & 2024 & & 47.44 & 69.36 & 48.48 & 70.23 & 50.18 & 72.24 & 48.79 & 70.68 & 59.74 \\
\,Re-ranking*~\cite{reranking} & BLIP & 2024 & & 48.14 & 71.43 & 50.15 & 71.25 & 55.23 & 76.80 & 51.17 & 73.13 & 62.15 \\
\,CaLa~\cite{cala} & BLIP-2 & 2024 & & 42.38 & 66.08 & 46.76 & 68.16 & 50.93 & 73.42 & 46.69 & 69.22 & 57.96 \\
\,SPRC~\cite{sprc} & BLIP-2 & 2024 & & 49.18 & 72.43 & 55.64 & 73.89 & 59.35 & 78.58 & 54.72 & 74.97 & 64.85 \\
\,SPRC†~\cite{spn} & BLIP-2 & 2024 & & 50.57 & \textbf{74.12} & 57.70 & 75.27 & 60.84 & \underline{79.96} & 56.37 & \underline{76.45} & 66.41 \\
\,CCIN~\cite{ccin} & BLIP-2 & 2025 & & 49.38 & 72.58 & 55.93 & 74.14 & 57.93 & 77.56 & 54.41 & 74.76 & 64.59 \\
\,CIR-LVLM~\cite{cir_lvlm} & MLLM & 2025 & & 50.42 & 73.57 & \textbf{58.59} & \underline{75.86} & 59.61 & 78.99 & 56.21 & 76.14 & 66.17 \\
\midrule
\rowcolor{Compositor}
\,\textbf{\textit{DetailFusion} (Ours)} & BLIP-2 & & & \underline{50.92} & 73.82 & 57.90 & 75.81 & \underline{60.94} & 79.60 & \underline{56.59} & 76.41 & \underline{66.50} \\
\rowcolor{Compositor}
\,\textbf{\textit{DetailFusion}† (Ours)} & BLIP-2 & & & \textbf{51.34} & \underline{74.05} & \underline{58.12} & \textbf{75.95} & \textbf{61.22} & \textbf{80.09} & \textbf{56.89} & \textbf{76.70} & \textbf{66.79} \\
\bottomrule
\end{tabular}
}
\end{large}
\end{center}
\vspace{-\baselineskip}
\end{table}

Table~\ref{tab:fiq_results_full} presents the complete results of our method compared to state-of-the-art approaches on the FashionIQ validation set. It is evident that, compared to SPRC~\cite{sprc}, our method achieves improvements of \textbf{+1.87} and \textbf{+1.44} in Average Recall@10 and 50, respectively. These results underscore the effectiveness of managing different levels of granularity through separate branches. Additionally, compared to CIR-LVLM~\cite{cir_lvlm}, which excels in understanding user intent within the fashion domain, our method achieves improvements of \textbf{+1.87} and \textbf{+1.44} in Average Recall@10 and 50, respectively, indicating that \textbf{\textit{DetailFusion}} enhances user intent understanding by improving detail capabilities.

Since the FashionIQ dataset does not include a retrieval subset composed of similar images, it is challenging to observe changes in the model's detail distinguish capabilities. Compared to methods using the same backbone, \textbf{\textit{DetailFusion}} achieves substantial improvements, demonstrating that the enhancement of detail distinguish capabilities also contributes effectively for CIR in specific domains.

However, due to the absence of a retrieval subset in the FashionIQ dataset, the SPN method described in Section~\textbf{\ref{sec:appendix_spn}} can not be implemented. Consequently, only the basic SPN~\cite{spn} method can be applied, resulting in negligible performance gains. We discuss that this limitation arises because the key component of the SPN lies in SN, which allows each query to encounter with all negative samples during the training process. This approach effectively enhances global information understanding, significantly improving the performance of the GM branch. However, it conflicts with our enhanced detail perception and processing capabilities, leading to suboptimal performance. This observation is consistent with the results obtained when applying the SPN method directly to the CIRR dataset.

\section{Additional Experiments and Ablation Studies}
\label{sec:appendix_experiment_ablation}


In this section, we conduct additional experimental analyses and ablation studies, including: The zero-shot capability of \textbf{\textit{DetailFusion}} and the cross-domain adaptability of its enhanced detail capabilities in Subsection~\ref{subsec:appendix_cross-domain_adaptability}; Quantitative analysis of computational complexity in Subsection~\ref{subsec:appendix_computational_complexity}; The impact of the number of cross-attention layers within the \textit{Adaptive Feature Compositor} in Subsection~\ref{subsec:appendix_cross-attention_layers}. Additionally, in Subsection~\ref{subsec:appendix_ablation_results}, we present the complete experimental results corresponding to the ablation studies discussed in Subsection~\ref{subsec:ablation_study}.

\subsection{Cross-Domain Adaptability of \textbf{\textit{DetailFusion}}}
\label{subsec:appendix_cross-domain_adaptability}

To investigate the cross-domain generalization ability of our proposed \textbf{\textit{DetailFusion}} framework and further validate the universality of the model's enhanced detail perception and processing capabilities in CIR, we conducted zero-shot CIR evaluations on the CIRCO~\cite{searle} and GeneCIS~\cite{genecis} test sets. The corresponding results are presented in Table~\ref{tab:circo_results} and Table~\ref{tab:genecis_results}, respectively.

\begin{table}[t]
\vspace{-0.5\baselineskip}
\caption{Zero-shot CIR performance on the CIRCO~\cite{searle} test set.}
\label{tab:circo_results}
\begin{center}
\renewcommand{\arraystretch}{1.05}
\resizebox{0.8\linewidth}{!}{
\begin{tabular}{lcc|cccc}
\toprule
\,\textbf{Method} & \textbf{Publication} & \textbf{Supervised} & $\text{mAP@5}$ & $\text{mAP@10}$ & $\text{mAP@25}$ & $\text{mAP@50}$ \\
\midrule
\,CompoDiff~\cite{compodiff} & TMLR 2024 & \textcolor{red}{\XSolidBrush} & 15.3 & 17.7 & 19.5 & 21.0 \\
\,LinCIR~\cite{lincir} & CVPR 2024 & \textcolor{red}{\XSolidBrush} & 19.7 & 21.0 & 23.1 & 24.2 \\
\,PrediCIR~\cite{predicir} & CVPR 2025 & \textcolor{red}{\XSolidBrush} & \textbf{23.7} & \textbf{24.6} & \textbf{25.4} & \textbf{26.0} \\
\midrule
\,Q-Former~\cite{blip2} & - & \textcolor{teal}{\Checkmark} & 17.5 & 19.2 & 21.0 & 22.3 \\
\,SPRC~\cite{sprc} & ICLR 2024 & \textcolor{teal}{\Checkmark} & 19.1 & 20.7 & 22.6 & 23.8 \\
\rowcolor{Compositor}
\multicolumn{2}{l}{\,\textbf{\textit{DetailFusion} (Ours)}} & \textcolor{teal}{\Checkmark} & \underline{22.8} & \underline{23.9} & \underline{24.6} & \underline{25.4} \\
\bottomrule
\end{tabular}
}
\end{center}
\end{table}

\begin{table}[h!]
\vspace{-0.5\baselineskip}
\caption{Zero-shot CIR performance on the GeneCIS~\cite{genecis} test set.}
\label{tab:genecis_results}
\begin{center}
\begin{large}
\renewcommand{\arraystretch}{1.15}
\resizebox{\linewidth}{!}{
\begin{tabular}{l|ccc|ccc|ccc|ccc|c}
\toprule
\multirow{2.4}{*}{\,\textbf{Method}} & \multicolumn{3}{c|}{\textbf{Focus Attribute}} & \multicolumn{3}{c|}{\textbf{Change Attribute}} & \multicolumn{3}{c|}{\textbf{Focus Object}} & \multicolumn{3}{c|}{\textbf{Change Object}} & \textbf{Avg.} \\
\cmidrule(lr){2-4} \cmidrule(lr){5-7} \cmidrule(lr){8-10} \cmidrule(lr){11-13} \cmidrule(lr){14-14}
 & R@1 & R@2 & R@3 & R@1 & R@2 & R@3 & R@1 & R@2 & R@3 & R@1 & R@2 & R@3 & R@1 \\
\midrule
\,CompoDiff~\cite{compodiff} & 14.3 & 26.7 & 38.4 & \underline{19.7} & 28.8 & 37.4 & 9.2 & 19.1 & 25.8 & 18.7 & 31.7 & 40.2 & 15.5 \\
\,LinCIR~\cite{lincir} & \underline{19.1} & \underline{33.0} & \underline{42.3} & 17.6 & \underline{30.2} & 38.1 & 10.1 & 19.1 & 28.1 & 7.9 & 16.3 & 25.7 & 13.7 \\
\,PrediCIR~\cite{predicir} & \textbf{19.3} & \textbf{33.2} & \textbf{42.7} & \textbf{19.9} & \textbf{30.7} & \underline{38.9} & \textbf{12.8} & 19.4 & \textbf{32.3} & 18.9 & \underline{32.2} & 40.6 & \textbf{18.7} \\
\midrule
\,Q-Former~\cite{blip2} & 14.6 & 26.0 & 37.7 & 16.4 & 29.1 & 38.2 & 10.4 & 20.4 & 29.6 & 18.7 & 31.3 & 41.5 & 15.0 \\
\,SPRC~\cite{sprc} & 14.9 & 27.7 & 39.0 & 16.7 & 28.2 & \underline{38.9} & 11.9 & \underline{22.8} & 31.6 & \underline{19.7} & \underline{32.2} & \underline{42.7} & 15.8 \\
\rowcolor{Compositor}
\,\textbf{\textit{DetailFusion} (Ours)} & 15.3 & 27.9 & 39.5 & 18.4 & 30.0 & \textbf{40.3} & \underline{12.5} & \textbf{23.6} & \underline{32.2} & \textbf{23.2} & \textbf{36.4} & \textbf{46.4} & \underline{17.3} \\
\bottomrule
\end{tabular}
}
\end{large}
\end{center}
\vspace{-\baselineskip}
\end{table}

All the supervised models were trained on the CIRR dataset prior to the evaluation on these unseen datasets for a fair comparison. It is evident that \textbf{\textit{DetailFusion}} surpasses SPRC~\cite{sprc} on both datasets and performs comparably to specially designed zero-shot CIR methods~\cite{lincir, predicir}, demonstrating that our approach exhibits superior zero-shot generalization performance among supervised CIR methods. Specifically, on the CIRCO test set, compared to SPRC, our method achieves improvements of \textbf{+3.7} and \textbf{+3.2} in mAP@5 and mAP@10, respectively. On the GeneCIS test set, compared to SPRC, our method achieves an improvement of \textbf{+1.5} in Average Recall@1 across the four subsets.

Furthermore, \textbf{\textit{DetailFusion}} significantly outperforms the single Q-Former~\cite{blip2} model, particularly on the ``Change'' metrics of GeneCIS test set, achieving improvements of \textbf{+2.0} and \textbf{+4.5} in Recall@1 for the Change Attribute and Change Object subsets, respectively. These findings further confirm the cross-domain generalization and robustness of our enhanced detail capabilities, which can be effectively transferred to any specific domain in CIR. Moreover, the experimental results underscores the pivotal role of atomic detail variation priors derived from the image editing dataset.

\subsection{Quantitative Analysis of Computational Complexity}
\label{subsec:appendix_computational_complexity}

\begin{table}[h!]
\vspace{-\baselineskip}
\caption{Comparison of \textbf{computational cost} on the CIRR and FashionIQ datasets.}
\label{tab:computational_cost}
\begin{center}
\begin{large}
\renewcommand{\arraystretch}{1.1}
\resizebox{0.75\linewidth}{!}{
\begin{tabular}{l|ccc|ccc}
\toprule
\multirow{2.5}{*}{\,\textbf{Method}} & \multicolumn{3}{c|}{\textbf{CIRR}} & \multicolumn{3}{c}{\textbf{FashionIQ}} \\
\cmidrule(lr){2-4} \cmidrule(lr){5-7}
 & Train $\downarrow$ & Inference $\downarrow$ & Avg.R $\uparrow$ & Train $\downarrow$ & Inference $\downarrow$ & Avg.R $\uparrow$ \\
\midrule
\,Re-ranking~\cite{reranking} & 45h & 26.01s & 81.78 & 34h & 54.38s & 62.15 \\
\,Q-Former~\cite{blip2} & 9.1h & 19.85s & 82.03 & 6.8h & 41.44s & 63.41 \\
\,SPRC~\cite{sprc} &  12.6h & 27.55s & 82.66 & 9.5h & 57.61s & 64.85 \\
\rowcolor{Compositor}
\,\textbf{\textit{DetailFusion} (Ours)} & 14.2h & 31.07s & 84.01 & 10.7h & 65.03s & 66.50 \\
\bottomrule
\end{tabular}
}
\end{large}
\end{center}
\vspace{-0.5\baselineskip}
\end{table}

We evaluate the computational costs of our proposed \textbf{\textit{DetailFusion}} in comparison with representative supervised CIR methods, including Re-ranking~\cite{reranking} and SPRC~\cite{sprc}, on the CIRR and FashionIQ datasets. Specifically, we consider the training time, inference time, and average recall as evaluation metrics, as shown in Table~\ref{tab:computational_cost}. All experiments were conducted on a single Tesla V100 GPU with 32GB of memory. The training time refers to the duration required for the model to converge to the optimum on the training set, while the inference time denotes the time for complete testing on the validation set.  According to Table~\ref{tab:computational_cost}, the overall computational cost during training for \textbf{\textit{DetailFusion}} is 10\%–20\% higher than that of single-stage supervised CIR methods. The inference time is approximately 1.5 times that of a single Q-Former and only 12\% slower than SPRC. Consequently, our method accepts a modest increase in computational overhead while achieving substantial improvements compared to SPRC, thereby realizing a better trade-off between accuracy and efficiency.

\subsection{Impact of the Number of Cross-Attention Layers in the Compositor}
\label{subsec:appendix_cross-attention_layers}

\begin{table}[t]
\vspace{-0.5\baselineskip}
\caption{Ablation of the \textbf{number of cross-attention layers} in Compositor on the CIRR validation set.}
\label{tab:ablation_number_layers_full}
\begin{center}
\begin{footnotesize}
\renewcommand{\arraystretch}{1.1}
\resizebox{0.75\linewidth}{!}{
\begin{tabular}{cc|cccc|ccc|c}
\toprule
\multirow{2.5}{*}{\textbf{\textit{M}}} & \multirow{2.5}{*}{\textbf{\textit{N}}} & \multicolumn{4}{c|}{\textbf{$\text{Recall@K}$}} & \multicolumn{3}{c|}{\textbf{$\text{Recall}_\text{subset}\text{@K}$}} & \multirow{2.5}{*}{\textbf{Avg.}} \\
\cmidrule(lr){3-6} \cmidrule(lr){7-9}
 & & \textbf{K=1} & \textbf{K=5} & \textbf{K=10} & \textbf{K=50} & \textbf{K=1} & \textbf{K=2} & \textbf{K=3} & \\
\midrule
\textbf{0} & \textbf{0} & 56.52 & 85.42 & 92.01 & 98.15 & 81.87 & 93.77 & 97.49 & 83.64 \\
\textbf{1} & \textbf{1} & 56.40 & 85.63 & 92.08 & 98.21 & 82.04 & 93.85 & 97.39 & 83.83 \\
\textbf{0} & \textbf{2} & 56.44 & 85.41 & 92.13 & 98.17 & 82.08 & 93.78 & 97.42 & 83.74 \\
\textbf{2} & \textbf{0} & 56.62 & 85.55 & 92.19 & 98.18 & 81.97 & 93.88 & 97.46 & 83.76 \\
\rowcolor{Compositor}
\textbf{2} & \textbf{2} & \textbf{56.83} & \textbf{85.75} & \textbf{92.37} & \textbf{98.23} & \textbf{82.28} & \textbf{94.07} & \textbf{97.56} & \textbf{84.01} \\
\textbf{4} & \textbf{4} & 56.47 & 85.53 & 92.09 & 98.11 & 82.21 & 93.95 & 97.51 & 83.87 \\
\bottomrule
\end{tabular}
}
\end{footnotesize}
\end{center}
\vspace{-\baselineskip}
\end{table}

In addition to employing the attention mechanism within the Feature Extraction Block to facilitate interactions between features from both branches and to inject fine-grained information, we evaluated the impact of varying the number of cross-attention layers in the \textit{Adaptive Feature Compositor}. Specifically, we examined the number of two types of cross-attention layers, denoted by $M$ and $N$, as illustrated in Fig.~\ref{fig:module_architecture}. The results are presented in Table~\ref{tab:ablation_number_layers_full}. Setting $M = 0$ implies the use of only same-side cross-attention. Under this configuration, the Compositor cannot perceive complementary information, resulting in inadequate guidance for fine-grained detail injection. Conversely, setting $N = 0$ restricts the model to use only cross-side cross-attention, preventing the Compositor from effectively refining and injecting useful fine-grained information. Both scenarios lead to a noticeable decline in retrieval performance. When $M = N$, we observed that using two cross-attention layers achieves optimal performance. This is because a single layer fails to enable sufficient interaction and precise information extraction, whereas an excessive number of layers increase the Compositor’s training difficulty and may exacerbate overfitting. Accordingly, we set $M = N = 2$ for the \textit{Adaptive Feature Compositor} to achieve optimal performance during training.

\subsection{Complete Ablation Experimental Results}
\label{subsec:appendix_ablation_results}

For ablation experiments that do not involve the Compositor or the third training stage, we sum the similarity scores of the query with all candidate images from both branches after joint fine-tuning and then sort them to obtain the retrieval results. This approach encompasses the first three rows of Table~\ref{tab:ablation_training_stage_full}, Table~\ref{tab:ablation_loss_function_full}, the first row of Table~\ref{tab:ablation_block_architecture_full}, Table~\ref{tab:ablation_trade_off_cirr}, and Table~\ref{tab:ablation_trade_off_fiq}.
Additionally, to eliminate the influence of the trade-off hyper-parameter, we set $\gamma = 1$ for experiments and investigate other factors.

\begin{table}[h!]
\vspace{-0.5\baselineskip}
\caption{Ablation of the \textbf{pre-training} and \textbf{dual-branch architecture} on the CIRR validation set.}
\label{tab:ablation_data_framework_full}
\begin{center}
\renewcommand{\arraystretch}{1.1}
\resizebox{0.95\linewidth}{!}{
\begin{tabular}{cc|cccc|ccc|c}
\toprule
\multirow{2.5}{*}{ \textbf{Method} } & \multirow{2.5}{*}{\textbf{Pre-training}} & \multicolumn{4}{c|}{\textbf{$\text{Recall@K}$}} & \multicolumn{3}{c|}{\textbf{$\text{Recall}_\text{subset}\text{@K}$}} & \multirow{2.5}{*}{\textbf{Avg.}} \\
\cmidrule(lr){3-6} \cmidrule(lr){7-9}
 & & \textbf{K=1} & \textbf{K=5} & \textbf{K=10} & \textbf{K=50} & \textbf{K=1} & \textbf{K=2} & \textbf{K=3} & \\
\midrule
SPRC~\cite{sprc} & \textcolor{red}{\XSolidBrush} & 54.32 & 84.32 & 90.98 & 97.98 & 81.00 & 93.02 & 97.16 & 82.66 \\
SPRC~\cite{sprc} & \textcolor{teal}{\Checkmark} & 54.57 & 84.43 & 91.09 & 98.01 & 81.35 & 93.17 & 97.26 & 82.89 \\
\rowcolor{GM}
Only \textit{GM} Branch & \textcolor{red}{\XSolidBrush} & 55.05 & 84.86 & 91.58 & 98.05 & 79.19 & 92.49 & 96.72 & 82.03 \\
\rowcolor{DI}
Only \textit{DI} Branch & \textcolor{teal}{\Checkmark} & 53.69 & 83.89 & 90.91 & 97.82 & 81.73 & 93.67 & 97.46 & 82.81 \\
\rowcolor{Compositor}
\textbf{\textit{DetailFusion}} & \textcolor{teal}{\Checkmark} & \textbf{56.83} & \textbf{85.75} & \textbf{92.37} & \textbf{98.23} & \textbf{82.28} & \textbf{94.07} & \textbf{97.56} & \textbf{84.01} \\
\bottomrule
\end{tabular}
}
\end{center}
\end{table}

\begin{table}[h!]
\vspace{-0.8\baselineskip}
\caption{Ablation of the \textbf{training strategies} for the framework on the CIRR validation set.}
\label{tab:ablation_training_stage_full}
\begin{center}
\begin{footnotesize}
\renewcommand{\arraystretch}{1.15}
\resizebox{0.85\linewidth}{!}{
\begin{tabular}{c|cccc|ccc|c}
\toprule
\multirow{2.4}{*}{\textbf{Training strategy}} & \multicolumn{4}{c|}{\textbf{$\text{Recall@K}$}} & \multicolumn{3}{c|}{\textbf{$\text{Recall}_\text{subset}\text{@K}$}} & \multirow{2.4}{*}{\textbf{Avg.}} \\
\cmidrule(lr){2-5} \cmidrule(lr){6-8}
 & \textbf{K=1} & \textbf{K=5} & \textbf{K=10} & \textbf{K=50} & \textbf{K=1} & \textbf{K=2} & \textbf{K=3} & \\
\midrule
Only Stage 2 & 55.03 & 84.88 & 92.01 & 98.04 & 80.72 & 93.28 & 97.03 & 82.80 \\
Hybrid of Stage 1\&2 & 54.84 & 84.81 & 91.72 & 97.92 & 81.25 & 93.45 & 97.25 & 83.03 \\
\rowcolor{gray!15}
Stage 1 \& Stage 2 & 55.92 & 85.29 & 92.07 & 98.18 & 81.63 & 93.71 & 97.44 & 83.46 \\
\rowcolor{Compositor}
\textbf{\textit{DetailFusion} (Full)} & \textbf{56.83} & \textbf{85.75} & \textbf{92.37} & \textbf{98.23} & \textbf{82.28} & \textbf{94.07} & \textbf{97.56} & \textbf{84.01} \\
\bottomrule
\end{tabular}
}
\end{footnotesize}
\end{center}
\end{table}

\begin{table}[h!]
\caption{Ablation of the \textbf{loss functions} for the DI branch on the CIRR validation set.}
\label{tab:ablation_loss_function_full}
\begin{center}
\begin{footnotesize}
\renewcommand{\arraystretch}{1.15}
\resizebox{0.85\linewidth}{!}{
\begin{tabular}{cc|cccc|ccc|c}
\toprule
\multirow{2.5}{*}{\textbf{Stage 1}} & \multirow{2.5}{*}{\textbf{Stage 2}} & \multicolumn{4}{c|}{\textbf{$\text{Recall@K}$}} & \multicolumn{3}{c|}{\textbf{$\text{Recall}_\text{subset}\text{@K}$}} & \multirow{2.5}{*}{\textbf{Avg.}} \\
\cmidrule(lr){3-6} \cmidrule(lr){7-9}
 & & \textbf{K=1} & \textbf{K=5} & \textbf{K=10} & \textbf{K=50} & \textbf{K=1} & \textbf{K=2} & \textbf{K=3} & \\
\midrule
\textcolor{red}{$\mathcal{L}_{G \! M}$} & \textcolor{red}{$\mathcal{L}_{G \! M}$} & 55.13 & 84.76 & 91.87 & 97.87 & 79.50 & 92.54 & 96.80 &	82.13 \\
\textcolor{teal}{$\mathcal{L}_{D \! I}$} & \textcolor{red}{$\mathcal{L}_{G \! M}$} & 55.47 & 85.05 & 91.89 & 98.02 & 80.24 & 92.78 & 97.08 &	82.65 \\
\textcolor{red}{$\mathcal{L}_{G \! M}$} & \textcolor{teal}{$\mathcal{L}_{D \! I}$} & 55.51 & 85.24 & 91.96 & 98.06 & 81.20 & 93.47 & 97.27 & 83.22 \\
\rowcolor{gray!15}
\textcolor{teal}{$\mathcal{L}_{D \! I}$} & \textcolor{teal}{$\mathcal{L}_{D \! I}$} & \textbf{55.92} & \textbf{85.29} & \textbf{92.07} & \textbf{98.18} & \textbf{81.63} & \textbf{93.71} & \textbf{97.44} & \textbf{83.46} \\
\bottomrule
\end{tabular}
}
\end{footnotesize}
\end{center}
\end{table}

\begin{table}[h!]
\vspace{-0.5\baselineskip}
\caption{Ablation of the \textbf{block architectures} for the Compositor on the CIRR validation set.}
\label{tab:ablation_block_architecture_full}
\begin{center}
\renewcommand{\arraystretch}{1.2}
\resizebox{0.9\linewidth}{!}{
\begin{tabular}{cc|cccc|ccc|c}
\toprule
\multirow{2.5}{*}{\textbf{Extraction}} & \multirow{2.5}{*}{\textbf{Fusion}} & \multicolumn{4}{c|}{\textbf{$\text{Recall@K}$}} & \multicolumn{3}{c|}{\textbf{$\text{Recall}_\text{subset}\text{@K}$}} & \multirow{2.5}{*}{\textbf{Avg.}} \\
\cmidrule(lr){3-6} \cmidrule(lr){7-9}
 & & \textbf{K=1} & \textbf{K=5} & \textbf{K=10} & \textbf{K=50} & \textbf{K=1} & \textbf{K=2} & \textbf{K=3} & \\
\midrule
\rowcolor{gray!15}
- & \textit{Average} & 55.92 & 85.29 & 92.07 & 98.18 & 81.63 & 93.71 & 97.44 & 83.46 \\
\textit{Concat} & \multirow{2}{*}{\textcolor{Dandelion}{\textbf{${M\!L\!P}_{\!\varphi}$}} \!\!\! \textbf{+} \!\! \textcolor{ForestGreen}{\textbf{${M\!L\!P}_{\!\psi}$}}} & 56.52 & 85.42 & 92.01 & 98.15 & 81.87 & 93.77 & 97.49 & 83.64 \\
\textit{Projection} & & 56.71 & 85.67 & 92.21 & 98.21 & 82.01 & 93.85 & 97.52 & 83.84 \\
\multirow{2}{*}{\textbf{\textit{Attention}}} & \textcolor{Dandelion}{\textbf{${M\!L\!P}_{\!\varphi}$}} & 55.85 & 85.19 & 91.96 & 98.06 & 81.75 & 93.76 & 97.46 & 83.47 \\
 & \textcolor{ForestGreen}{\textbf{${M\!L\!P}_{\!\psi}$}} & 55.56 & 85.17 & 91.92 & 98.01 & 81.39 & 93.33 & 97.35 & 83.28 \\
\rowcolor{Compositor}
\textbf{\textit{Attention}} & \textcolor{Dandelion}{\textbf{${M\!L\!P}_{\!\varphi}$}} \!\!\! \textbf{+} \!\! \textcolor{ForestGreen}{\textbf{${M\!L\!P}_{\!\psi}$}} & \textbf{56.83} & \textbf{85.75} & \textbf{92.37} & \textbf{98.23} & \textbf{82.28} & \textbf{94.07} & \textbf{97.56} & \textbf{84.01} \\
\bottomrule
\end{tabular}
}
\end{center}
\end{table}

\begin{table}[h!]
\vspace{-0.5\baselineskip}
\caption{Ablation of the \textbf{values of trade-off hyper-parameter $\mathbf{\gamma}$} on the CIRR validation set.}
\label{tab:ablation_trade_off_cirr}
\begin{center}
\begin{footnotesize}
\renewcommand{\arraystretch}{1.15}
\resizebox{0.7\linewidth}{!}{
\begin{tabular}{c|cccc|ccc|c}
\toprule
\multirow{2.4}{*}{\,\,\,$\mathbf{\gamma}$\,\,\,} & \multicolumn{4}{c|}{\textbf{$\text{Recall@K}$}} & \multicolumn{3}{c|}{\textbf{$\text{Recall}_\text{subset}\text{@K}$}} & \multirow{2.4}{*}{\textbf{Avg.}} \\
\cmidrule(lr){2-5} \cmidrule(lr){6-8}
 & \textbf{K=1} & \textbf{K=5} & \textbf{K=10} & \textbf{K=50} & \textbf{K=1} & \textbf{K=2} & \textbf{K=3} & \\
\midrule
\textbf{0.5} & 55.27 & 85.10 & 91.89 & 98.05 & 81.30 & 93.28 & 97.23 & 83.20 \\
\rowcolor{gray!15}
\textbf{1.0} & 55.92 & 85.29 & 92.07 & 98.18 & 81.63 & 93.71 & 97.44 & 83.46 \\
\textbf{1.5} & 56.04 & 85.67 & 92.30 & 98.15 & 81.73 & 93.69 & 97.42 & 83.70 \\
\rowcolor{Compositor}
\textbf{2.0} & \textbf{56.21} & \textbf{85.89} & \textbf{92.44} & \textbf{98.25} & \textbf{81.91} & \textbf{93.88} & \textbf{97.61} & \textbf{83.90} \\
\textbf{2.5} & 55.94 & 85.75 & 92.42 & 98.19 & 81.70 & 93.64 & 97.51 & 83.72 \\
\textbf{3.0} & 55.49 & 85.51 & 91.92 & 98.11 & 81.56 & 93.33 & 97.36 & 83.53 \\
\bottomrule
\end{tabular}
}
\end{footnotesize}
\end{center}
\end{table}

\begin{table}[h!]
\vspace{-0.5\baselineskip}
\caption{Ablation of the \textbf{values of trade-off hyper-parameter $\mathbf{\gamma}$} on the FashionIQ validation set.}
\label{tab:ablation_trade_off_fiq}
\begin{center}
\renewcommand{\arraystretch}{1.2}
\resizebox{0.8\linewidth}{!}{
 \begin{tabular}{c|cc|cc|cc|cc|c}
\toprule
\multirow{2.7}{*}{$\mathbf{\gamma}$} & \multicolumn{2}{c|}{\textbf{Dress}} & \multicolumn{2}{c|}{\textbf{Shirt}} & \multicolumn{2}{c|}{\textbf{Toptee}} & \multicolumn{2}{c|}{\textbf{Average}} & \multirow{2.7}{*}{\textbf{Avg.}} \\
\cmidrule(lr){2-3} \cmidrule(lr){4-5} \cmidrule(lr){6-7} \cmidrule(lr){8-9}
 & \textbf{R@10} & \textbf{R@50} & \textbf{R@10} & \textbf{R@50} & \textbf{R@10} & \textbf{R@50} & \textbf{R@10} & \textbf{R@50} & \\
\midrule
\textbf{0.5} & 50.57 & 72.63 & 55.84 & \underline{75.12} & 59.92 & 78.89 & 55.44 & 75.55 & 65.50 \\
\rowcolor{gray!15}
\textbf{1.0} & 50.37 & 72.68 & 56.48 & 74.98 & \textbf{59.97} & 79.45 & \underline{55.61} & 75.70 & 65.65 \\
\rowcolor{Compositor}
\textbf{1.5} & \textbf{50.76} & \textbf{72.95} & \underline{56.50} & \textbf{75.24} & \underline{59.96} & \textbf{80.04} & \textbf{55.74} & \textbf{76.08} & \textbf{65.91} \\
\textbf{2.0} & \underline{50.72} & \underline{72.88} & 56.23 & 74.58 & 59.77 & \underline{79.91} & 55.57 & \underline{75.79} & \underline{65.68} \\
\textbf{2.5} & 50.02 & 72.63 & \textbf{57.21} & 74.78 & 59.51 & 79.70 & 55.58 & 75.71 & 65.64 \\
\textbf{3.0} & 50.47 & 72.63 & 56.08 & 74.93 & 59.71 & 79.50 & 55.42 & 75.69 & 65.55 \\
\bottomrule
\end{tabular}
}
\end{center}
\end{table}

\section{Additional Qualitative Comparison and Analysis}
\label{sec:appendix_qualitative_comparison}

\begin{figure}[t]
\vspace{-0.5\baselineskip}
\begin{center}
\centerline{\includegraphics[width=1.02\linewidth, trim=0.3cm 0.4cm 0.6cm 0.2cm, clip]{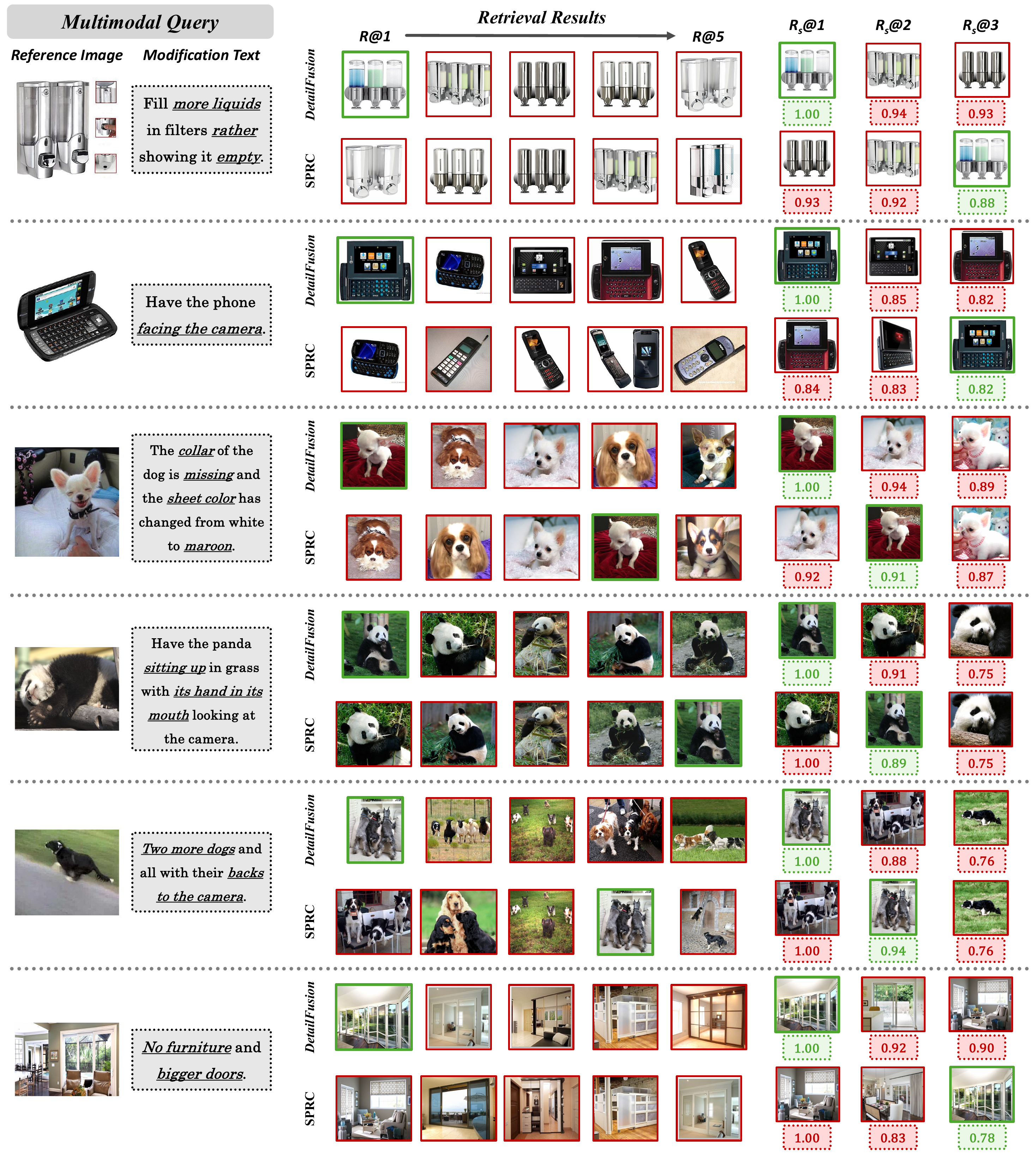}}
\caption{\textbf{Qualitative comparison of our method and SPRC on the CIRR validation set.} Images are arranged in descending order from left to right based on similarity to the multimodal query. The \textbf{\textcolor[HTML]{70ad47}{green boxes}} highlight the target image, while all non-target images are marked with \textbf{\textcolor[HTML]{aa0d06}{red boxes}}. In the subset retrieval results, we report the \textit{relative similarity score}, which defined as the normalized similarity between each retrieved image and the multimodal query, relative to all images in the gallery.}
\vspace{-1.5\baselineskip}
\label{fig:qualitative_comparison_sprc}
\end{center}
\end{figure}

As illustrated in Fig.~\ref{fig:qualitative_comparison_sprc}, our proposed \textbf{\textit{DetailFusion}} outperforms SPRC~\cite{sprc} in handling subtle image alterations and complex modification requirements in CIR, resulting in more accurate retrieval outcomes. According to the relative similarity scores reported in the subset retrieval results, our method excels at differentiating highly semantically similar images, as demonstrated by the clear margin between the correctly retrieved target image and the most similar image. In contrast, SPRC struggles to retrieve the target image among such confusing images, as these images exhibit closely relative similarity scores. Some retrieval failure cases are explained and analyzed in Section~\ref{sec:appendix_limitation_future_work}.

\section{Limitation and Future Work}
\label{sec:appendix_limitation_future_work}

Despite the promising performance of the proposed \textbf{\textit{DetailFusion}} for supervised CIR, several limitations remain. Due to the use of the frozen ViT-G/14 from EVA-CLIP~\cite{eva_clip} as the visual encoder and the BLIP-2 pre-trained Q-Former~\cite{blip2} as the backbone, the model’s initialization heavily relies on the capabilities of these pre-trained models. This also implies that the inherent vision-language alignment issues introduced by these models are challenging to address effectively within our approach. For instance, accurately associating text descriptions of spatial orientations with their corresponding visual representations remains difficult. In future work, we plan to tackle these challenges by constructing more targeted, high-quality training data and incorporating additional validation modules to mitigate the limitations introduced by these pre-trained models.

\end{document}